\documentclass[conference]{IEEEtran}
\IEEEoverridecommandlockouts

\usepackage{cite}
\usepackage{amsmath,amssymb,amsfonts}
\usepackage{algorithm}
\usepackage{algpseudocodex}
\usepackage{graphicx}
\usepackage{textcomp}
\usepackage{xcolor}
\usepackage{subcaption}
\usepackage{tabularx}
\usepackage{booktabs}
\usepackage{multirow}
\usepackage{enumitem}
\usepackage{xurl}
\usepackage{hyperref}
\usepackage{cleveref}
\usepackage{marvosym}
\usepackage{threeparttable}

\algrenewcommand\algorithmiccomment[1]{%
  \hfill{\normalsize\itshape\color{gray}{\(\triangleright\)~#1}}%
}

\newcommand{\name}{\emph{UniAlign}}
\newcommand{\emphtitle}[1]{%
    \par\noindent
    {\normalfont\bfseries #1}
}

\crefname{section}{\S}{\S\S}
\Crefname{section}{\S}{\S\S}
\crefname{table}{Table}{Tables}
\Crefname{table}{Table}{Tables}
\crefname{figure}{Figure}{Figures}
\Crefname{figure}{Figure}{Figures}
\crefname{equation}{eq.}{eqs.}
\Crefname{equation}{Equation}{Equations}
\crefname{algorithm}{Algorithm}{Algorithms}
\Crefname{algorithm}{Algorithm}{Algorithms}

\makeatletter
\crefname{ALG@line}{line}{lines}
\Crefname{ALG@line}{Line}{Lines}
\makeatother

\def\BibTeX{{\rm B\kern-.05em{\sc i\kern-.025em b}\kern-.08em
    T\kern-.1667em\lower.7ex\hbox{E}\kern-.125emX}}
\begin{document}

\title{\name: A Model-Agnostic Framework for Robust Network Traffic Classification under Distribution Shifts
}

\author{
\IEEEauthorblockN{Tongze Wang\textsuperscript{\textdagger}, Xiaohui Xie\textsuperscript{\textdaggerdbl\,\Letter} \thanks{\textsuperscript{\Letter}Corresponding Authors: Xiaohui Xie (xiexiaohui@tsinghua.edu.cn) and Yong Cui (cuiyong@tsinghua.edu.cn)}, Wenduo Wang\textsuperscript{\textdaggerdbl}, Chuyi Wang\textsuperscript{\textdaggerdbl}, Yong Cui\textsuperscript{\textdaggerdbl\,\Letter}}
\IEEEauthorblockA{\textsuperscript{\textdagger}\textit{Institute for Network Sciences and Cyberspace, Tsinghua University}\\
\textsuperscript{\textdaggerdbl}\textit{Department of Computer Science and Technology, Tsinghua University}}
}

\maketitle

\begin{abstract}
Network traffic classification (NTC) models often suffer severe performance degradation when deployed in real-world environments due to distribution shifts caused by changing network conditions. Existing robustness-enhancing approaches are commonly coupled to specific model architectures or data settings, fail to generalize to state-of-the-art raw-byte–based NTC models, or incur significant training overhead. In this paper, we propose \name{}, a novel model-agnostic framework that improves the robustness of deep learning–based NTC models under distribution shifts. \name{} combines \emph{domain alignment fine-tuning}, which encourages the learning of domain-invariant traffic representations across heterogeneous network conditions, with \emph{stable model ensembling}, which enhances inference robustness by aggregating checkpoints within a flat loss region. The framework can be seamlessly integrated into existing supervised NTC models without requiring specific feature modalities or introducing non-constant additional training costs. We evaluate \name{} on three public datasets covering diverse distribution shifts, including encryption schemes, data collection devices, and attack behaviors. Experimental results on two representative NTC models demonstrate that, compared with standard training, \name{} improves average classification accuracy by 2.51\% and average F1 score by 2.71\%, outperforming the strongest baseline by 1.45\% in accuracy and 1.69\% in F1 score, while requiring only 12.4\%--53.9\% of the training time of all NTC-specific baselines.
\end{abstract}


\section{Introduction}

Network traffic classification~(NTC), which aims to identify the category of traffic generated by various applications or detect potential threats within communication behaviors, is crucial for several key areas: improving service quality \cite{eswara2019streaming, peng2024ptu}, safeguarding network security\cite{fu2021realtime, barradas2021flowlens, apruzzese2018effectiveness}, and facilitating compliance monitoring\cite{oh2023appsniffer, xue2024fingerprinting}. With the widespread adoption of encrypted protocols, there has been a significant proliferation of AI-based NTC models\cite{hayes2016k, taylor2017robust, liu2019fs, van2020flowprint, lotfollahi2020deep, lin2022bert, zhao2023yet, wang2024netmamba, zhou2025trafficformer, yang2025mm4flow, zhao2025sweet} proposed to accurately capture complex traffic communication patterns.

Owing to the superior modeling capabilities inherent in supervised machine learning and deep learning, these models typically exhibit excellent classification performance during laboratory evaluation, where the training and test data are often assumed to be independent and identically distributed (i.i.d.). However, this ideal data collection assumption rarely holds in real-world deployment environments. Crucially, due to dynamic network conditions-stemming from transmission path alterations, variations in communication protocols, or software upgrades-the feature distribution of traffic collected in the deployment environment frequently shifts from that of the training data. This distribution shift ultimately results in a marked degradation of model performance. 

Existing efforts to address distribution shifts in NTC can be broadly categorized into two groups. 
(1) \emph{Model-specific methods} aim to improve generalization by retraining subnetworks with newly collected data~\cite{chen2025cd}, filtering stable graph features~\cite{cui2025fg}, or exploiting the inherent generalization ability of large language models~\cite{lin2025respond}. However, these approaches are tightly coupled with specific data acquisition settings or model architectures, which limits their applicability to other NTC models.
(2) \emph{Model-agnostic methods} seek to enhance existing models using techniques such as data augmentation~\cite{bahramali2023realistic, xie2023rosetta} or meta learning~\cite{qing2025training}. 
Nevertheless, data augmentation techniques are typically constrained to models that rely on packet size traces, and therefore fail to improve pre-trained NTC models based on raw-byte representations, which are widely recognized as state-of-the-art in modern NTC~\cite{lin2022bert, zhao2023yet, wang2024netmamba, zhou2025trafficformer}. In addition, meta-learning-based approaches often suffer from suboptimal performance due to limited simulation of real-world shift scenarios, as well as substantial training overhead that scales linearly with the number of sub-tasks.

To overcome the above limitations, we propose \name{}, a novel model-agnostic framework designed to enhance the robustness of deep learning-based NTC models under distribution shifts. Our motivation is twofold. 
First, traffic generated by the same network behavior (e.g., web browsing or attack execution) tends to share intrinsic characteristics that remain consistent across different distributions. For instance, attack traffic generally exhibits abnormal patterns in terms of transmission rate, traffic volume, or connection completeness compared with benign traffic. 
This observation motivates us to improve NTC performance by learning cross-domain stable feature representations, which can seamlessly adapt to models with arbitrary input modalities while incurring no additional training complexity.
Second, selecting a single optimal checkpoint for inference is often insufficient for robust generalization. Under distribution shifts, the checkpoint that performs best on validation data may correspond to a sharp minimum, which is known to suffer more severe performance degradation than flatter minima on unseen data. Therefore, we aim to improve generalization by ensembling model checkpoints within a flat loss region.

Specifically, \name{} consists of two key modules: \emph{domain alignment fine-tuning} and \emph{stable model ensembling}. 
For both modules, we consider distribution shifts caused by different network conditions during traffic generation, transmission, and collection, where each condition is treated as a distinct domain. By constructing training and test sets with non-overlapping domains, we simulate the discrepancy between training and deployment environments. 

The challenge of domain alignment fine-tuning lies in identifying a comprehensive alignment target and balancing the relative scales of the alignment loss and the classification loss. To address this challenge, we first encourage models to learn domain-invariant representations by minimizing discrepancies in both first-order and second-order feature statistics across domains, and then increase the difficulty of the classification task by introducing label smoothing.

The challenge of stable model ensembling lies in identifying an appropriate flat loss region in an online manner and ensuring the stability of checkpoint averaging. To this end, we first locate a flat region in the validation loss trajectory guided by a heuristic rule, and then aggregate the corresponding model parameters using adaptive weights. By integrating both components, \name{} improves the generalization of NTC models across arbitrary feature modalities, without increasing training complexity compared with standard training procedures.
A comparison between \name{} and other model-agnostic robust training frameworks is summarized in \Cref{tab:method_cmp}.

We evaluate \name{} on three public datasets that cover diverse types of distribution shifts, including encryption scheme shifts, data collection device shifts, and attack behavior shifts. Compared with three robust training baselines specifically designed for NTC and four general-purpose robust learning baselines, all evaluated on two representative NTC models, \name{} consistently achieves the best performance under both i.i.d. and distribution-shifted settings. The main contributions of this paper are summarized as follows:
\begin{enumerate}[label=(\arabic*), left=0pt, labelsep=1em, itemsep=0pt, topsep=0pt]
    \item We propose \name{}, a novel model-agnostic robust training framework that improves the performance of supervised deep learning-based NTC models under distribution shifts.
    \item Our framework enables NTC models to learn stable traffic representations by minimizing cross-domain discrepancies and to select robust model parameters through filtered checkpoint averaging. These two components jointly improve generalization without requiring specific feature modalities or incurring non-constant additional training costs.
    \item We implement \name{} and conduct extensive experiments on three public datasets and two representative NTC models. Compared with standard training, our framework improves average classification accuracy by 2.51\% and average F1 score by 2.71\%, outperforming the strongest baseline by 1.45\% in average accuracy and 1.69\% in average F1 score, while requiring only 12.4\%--53.9\% of the training time of all NTC-specific baselines.
\end{enumerate}


\begin{table*}[htbp]
    \footnotesize
    \centering
    \begin{threeparttable}
    \begin{tabular}{c|c|c|c} \toprule
         Framework & Supported Modality & Extra Training Stage & Per-epoch Steps \\ \midrule
         Rosetta~\cite{xie2023rosetta} & Packet size traces only & Contrastive training + TIE module\,$^{\dagger}$ & $\mathcal{O}(1)$ \\ \midrule
         NetAugment~\cite{bahramali2023realistic} & Packet size traces only & Contrastive training & $\mathcal{O}(1)$ \\ \midrule
         MetaTraffic~\cite{qing2025training} & Arbitrary modality & None & $\mathcal{O}(\mathcal{T})$\,$^{*}$ \\ \midrule
         \name{} & Arbitrary modality & None & $\mathcal{O}(1)$ \\
         \bottomrule
    \end{tabular}
    \begin{tablenotes}
        \item $^{\dagger}$ Rosetta inserts an additional TIE module during fine-tuning, raising the per-step cost.
        \item $^{*}$ $\mathcal{T}$ denotes the number of meta-learning sub-tasks.
    \end{tablenotes}
    \end{threeparttable}
    \caption{Comparison of NTC-specific robust training frameworks.}
    \label{tab:method_cmp}
\end{table*}
\section{Threat Model}

This work considers distribution shift as an environmental threat to supervised deep learning–based NTC systems. Distribution shift arises when the statistical properties of traffic observed during deployment differ from those encountered during model training, leading to significant performance degradation.
Formally, let $\mathcal{X}$ denote the input feature space and $\mathcal{Y}$ the label space. The training dataset $D_{\text{tr}}$ and the test dataset $D_{\text{te}}$ are drawn from different joint distributions over $\mathcal{X} \times \mathcal{Y}$, denoted as $P_{\text{tr}}$ and $P_{\text{te}}$, respectively, where $P_{\text{tr}} \neq P_{\text{te}}$. We assume that labels remain consistent across environments (i.e., the label space does not change), while the input distribution and feature–label relationships may vary.

Our framework aims to enhance the robustness of supervised NTC models under such distribution shifts, without requiring access to test-time data or retraining on target domains. We adopt a threat model aligned with the standard domain generalization (DG) setting \cite{zhou2022domain}, which captures realistic deployment scenarios in operational networks.

\emphtitle{Domains and Data Organization.} We attribute distribution shifts primarily to dynamic and heterogeneous network conditions, rather than to adaptive adversaries who manipulate traffic with knowledge of the classifier. To simulate these conditions, traffic data are collected from multiple network environments, each of which is treated as a distinct \emph{domain}. Specifically:
\begin{itemize}[left=0pt, labelsep=1em, itemsep=0pt, topsep=0pt]
    \item The training data $D_{\text{tr}}$ consist of $S$ source domains, denoted as $D_{\text{tr}} = \{ D_{\text{tr}}^{i} \}_{i=1}^{S}$, where each domain $D_{\text{tr}}^{i}$ is sampled from a joint distribution $P_{\text{tr}}^{i}$ over $\mathcal{X} \times \mathcal{Y}$.
    \item The test data $D_{\text{te}}$ consist of $T$ target domains, denoted as $D_{\text{te}} = \{ D_{\text{te}}^{j} \}_{j=1}^{T}$, where each domain $D_{\text{te}}^{j}$ follows a joint distribution $P_{\text{te}}^{j}$ over $\mathcal{X} \times \mathcal{Y}$.
\end{itemize}
To ensure a non-trivial evaluation of robustness, we enforce that $P_{\text{tr}}^{i} \neq P_{\text{te}}^{j}, \ \forall i \in \{ 1, \dots, S\}, \ \forall j \in \{ 1, \dots, T\}$, which guarantees that target-domain traffic is strictly out-of-distribution with respect to all source domains. We further assume $S > 1$ for all considered settings, as multi-source domains are required by many robust training and domain generalization techniques.

\emphtitle{Considered Distribution Shift Types.}
We focus on three representative and practically relevant sources of distribution shift in real-world networks:

\begin{itemize}[left=0pt, labelsep=1em, itemsep=0pt, topsep=0pt]
    \item \emph{Encryption Scheme Shift.}
    The choice of TLS cipher suites affects both the plaintext fields of TLS handshake messages and the resulting traffic fingerprints. Since cipher suites are negotiated through \texttt{ClientHello} and \texttt{ServerHello} messages, changes in encryption configurations alter handshake structures, plaintext payload fields, and packet size distributions. These variations induce noticeable shifts in traffic features used by NTC models.
    \item \emph{Collection Device Shift.}  
    Traffic captured from different end hosts exhibits distributional differences due to variations in hardware platforms, operating systems, and protocol implementations. For example, older operating systems (e.g., Android 7) typically default to TLS 1.2, whereas newer systems (e.g., Android 10) adopt TLS 1.3, resulting in fewer handshake rounds and significantly different handshake message lengths. Such discrepancies lead to systematic shifts in traffic representations.
    \item \emph{Attack Behavior Shift.}  
    Network attacks manifest diverse traffic patterns depending on their objectives and execution strategies. For instance, port scanning attacks generate numerous short-lived connections across multiple ports, Distributed Denial-of-Service (DDoS) attacks cause sustained high-volume traffic surges, and web-based attacks often embed malicious payloads within HTTP requests or URLs. These behavioral differences introduce substantial variability in flow-level and packet-level features.
\end{itemize}

\section{Design}

\subsection{Overview}

\begin{figure*}[tp]
    \centering
    \includegraphics[scale=0.6]{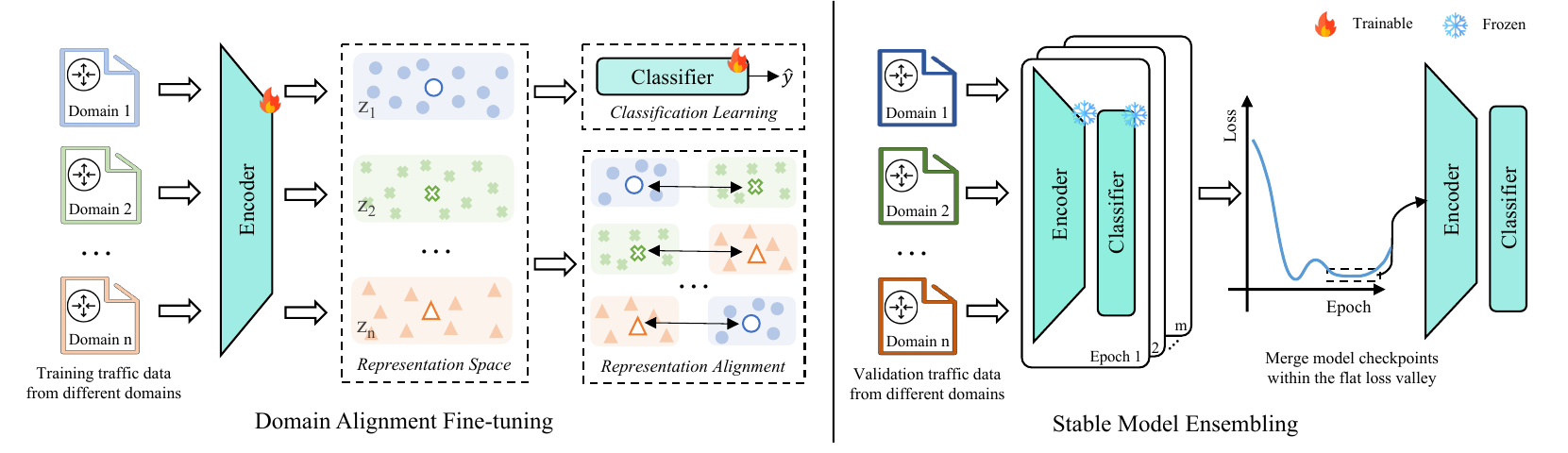}
    \caption{Overview of the \name{} framework. The framework consists of two modules: (left) a domain alignment fine-tuning module that incorporates an additional representation alignment loss to minimize cross-domain feature discrepancies, and (right) a stable model ensembling module that merges model checkpoints located in an identified flat loss valley.}
    \label{fig:system-design}
\end{figure*}

To achieve robust network traffic classification (NTC) under distribution shifts, \name{} adopts a two-stage strategy. First, it guides NTC models to learn domain-invariant traffic representations through \emph{domain alignment fine-tuning}, which minimizes representation discrepancies across domains. Second, it further enhances generalization via \emph{stable model ensembling}, which aggregates model weights located within a flat loss valley. The overall fine-tuning and inference workflow of \name{} is illustrated in \Cref{fig:system-design}.

Formally, an NTC model consists of an encoder $\mathcal{E}$ and a classifier $\mathcal{C}$, with parameters collectively denoted as $\theta$. The encoder $\mathcal{E}$ processes network traffic features in various input formats, including packet size sequences~\cite{liu2019fs, sirinam2018deep, deng2023robust}, byte tokens~\cite{lin2022bert, peng2024ptu, zhou2025trafficformer, yang2025mm4flow}, byte patches~\cite{zhao2023yet, hang2023flow, wang2024netmamba}, and byte graphs~\cite{zhang2023tfe}. It projects these features into initial embeddings and captures cross-feature dependencies using stacked deep neural network (DNN) blocks such as CNN, RNN, Transformer, and Mamba modules. After representation learning, the encoder outputs a feature vector $z_i = \mathcal{E}(x_i;\theta) \in \mathbb{R}^{D}$ for each input sample $x_i$, which is then fed into the classifier $\mathcal{C}$ to predict the corresponding traffic label.

During fine-tuning, domain alignment is implemented under a multi-task learning framework that jointly optimizes a classification learning module and a representation alignment module.
The classification module aims to model the traffic label distribution across all domains, thereby establishing fundamental classification capability. 
In parallel, the representation alignment module encourages the learning of domain-invariant features by minimizing representation discrepancies among different domains, which has been shown to improve NTC performance under distribution shifts~\cite{sun2016deep, li2018domain}.

During inference, \name{} further improves generalization through stable model ensembling, which enhances robustness by averaging model weights. Specifically, this procedure first identifies a flat region in the validation loss trajectory and then implicitly ensembles all models within this region by merging their corresponding weights. By aggregating weights from a flat loss valley, \name{} avoids selecting models located at sharp minima, which are known to be more sensitive to distribution shifts. As a result, this strategy yields improved robustness and generalization~\cite{foret2020sharpness, cha2021swad, zhuang2022surrogate}.

\subsection{Domain Alignment Fine-tuning}
Under independent and identically distributed (i.i.d.) assumptions, the standard fine-tuning procedure for NTC models optimizes only the classification loss (e.g., cross-entropy), implicitly assuming consistent feature distributions between training and test data. Such an assumption often breaks under real-world deployment, where traffic distributions vary across domains. To improve NTC generalization under distribution shifts, we introduce an additional optimization objective that explicitly enforces cross-domain representation alignment and fine-tune the NTC model within a multi-task learning framework.

\emphtitle{Representation Alignment.}
The key to robust NTC performance on unseen domains lies in learning feature representations that remain stable across multiple observed training domains. Existing approaches for this goal can be broadly categorized into three classes: meta learning, feature disentanglement, and representation alignment. 

Meta learning methods~\cite{finn2017model, li2018learning, qing2025training} construct multiple training tasks by randomly partitioning the training data into meta-train and meta-test sets. Model updates are then guided by gradients computed on the meta-test set using a temporary model trained on the corresponding meta-train set. Although meta learning benefits from incorporating signals from simulated unseen domains, it often yields limited generalization gains while incurring substantial training overhead, as the number of training iterations grows with the number of meta-tasks.
Feature disentanglement methods~\cite{ganin2016domain, li2018domain, nam2021reducing, ding2022domain} aim to separate domain-invariant and domain-specific components in the representation space via adversarial training on domain prediction. In practice, however, accurately assigning domain labels for network traffic data is challenging. An unreliable domain discriminator may therefore weaken the adversarial signal, leading to suboptimal disentanglement and degraded performance.
Representation alignment methods~\cite{sun2016deep, li2018domainmmd} instead suppress domain-specific information by directly minimizing cross-domain representation discrepancies, typically using distance metrics based on second-order statistics or reproducing kernel Hilbert spaces. These methods are computationally efficient and easy to implement, yet leave room for further improvement.

Motivated by these observations, we adopt an efficient and effective representation alignment strategy that jointly measures cross-domain distances using both first-order (mean) and second-order (covariance) statistics in the learned representation space.

Assume that the feature representation $z_i$ of a traffic sample $x_i$ can be decomposed into a domain-invariant component $z_i^{c}$ and a domain-specific component $z_i^{s}$, i.e., $z_i = (z_i^{c}, z_i^{s})$. By minimizing distances (e.g., Euclidean distance) between representations drawn from different domains, the NTC encoder is encouraged to focus on domain-invariant features.
However, directly enforcing instance-level pairwise alignment suffers from several limitations: (1) high computational complexity of $\mathcal{O}(N^2S^2)$, where $S$ denotes the number of training domains and $N$ the number of samples per domain; (2) overly restrictive constraints that may force samples from different classes but different domains to become artificially close; (3) sensitivity to noise introduced by outlier samples.

In contrast, distribution-level alignment based on sample statistics provides a softer and more robust constraint, while reducing computational complexity to $\mathcal{O}(NS^2)$. We first align first-order statistics by minimizing the discrepancy between domain-wise mean representations. The corresponding mean alignment loss is defined as:
\begin{equation}
    \begin{aligned}
        \mathcal{L}_{\text{mean}} &= \frac{S(S-1)}{2} \cdot \frac{1}{D} \sum_{i=1}^{S}\sum_{j=i+1}^{S} \| \mu_i - \mu_j \|_F^2 \\
        \mu_i &= \frac{1}{N} \sum_{k=1}^{N} z_{ik} \\
        \mu_j &= \frac{1}{N} \sum_{k=1}^{N} z_{jk}
    \end{aligned}
\end{equation}
where $D$ denotes the representation dimension, $z_{ik}$ is the feature representation of the $k$-th sample from the $i$-th training domain, and $\| \cdot \|_F^2$ represents the squared Frobenius norm.

While mean alignment mitigates first-order distribution shifts, it ignores variations in feature dispersion and inter-dimensional correlations. To address this limitation, we further incorporate covariance alignment to reduce second-order discrepancies across domains, thereby encouraging geometrically consistent representations while preserving discriminative structure. The covariance alignment loss is defined as:
\begin{equation}
    \begin{aligned}
        \mathcal{L}_{\text{cov}} &= \frac{S(S-1)}{2} \cdot \frac{1}{D^2} \sum_{i=1}^{S}\sum_{j=i+1}^{S} \| C_i - C_j \|_F^2 \\
        C_i &= \frac{1}{N-1} (Z_i - \mu_i)^{\top}(Z_i - \mu_i) \\
        C_j &= \frac{1}{N-1} (Z_j - \mu_j)^{\top}(Z_j - \mu_j)
    \end{aligned}
\end{equation}
where $Z_i \in \mathbb{R}^{N \times D}$ denotes the representation matrix of samples from the $i$-th training domain.

Finally, the overall representation alignment loss that jointly accounts for first-order and second-order statistics is given by:
\begin{equation}
    \mathcal{L}_{\text{align}} = \mathcal{L}_{\text{mean}} + \mathcal{L}_{\text{cov}}
\end{equation}

\emphtitle{Classification Learning.}
To train a supervised NTC model that effectively fits the training data distribution, the standard choice is the cross-entropy loss, which is equivalent to minimizing the Kullback–Leibler (KL) divergence between the ground-truth one-hot label distribution $y^{\text{one-hot}} = \delta_y$ and the predicted probability distribution $p$:

\begin{equation}
    \label{eq:ce}
    \mathcal{L}_{\text{CE}} = - \sum_{i=1}^{K} y^{\text{one-hot}}_i \log p_i
\end{equation}
where $K$ denotes the number of traffic classes.

Under the multi-task learning setting, it is necessary to balance the relative scales of the classification loss and the representation alignment loss. Otherwise, the model may overfit the task associated with a larger loss magnitude while under-optimizing the other.
To address this issue, several adaptive loss weighting strategies have been proposed. For instance, homoscedastic uncertainty~\cite{kendall2018multi} assigns each task a learnable noise parameter, while GradNorm\cite{chen2018gradnorm} dynamically balances heterogeneous objectives by equalizing their gradient magnitudes. Despite their effectiveness, these methods introduce additional parameters and optimization objectives, increasing training complexity and instability.

Empirically, we observe that the imbalance between task losses primarily stems from differences in task difficulty. For the classification task, NTC models can readily capture discriminative patterns, leading to a rapid decrease in loss magnitude. In contrast, domain alignment is inherently more challenging due to factors such as weakly discriminative domain characteristics in network traffic and label distribution mismatches across domain-specific mini-batches. A simple yet effective way to mitigate this imbalance is to increase the difficulty of the classification task.

Following prior works on label smoothing~\cite{szegedy2016rethinking, pereyra2017regularizing, muller2019does}, we replace the hard one-hot label $y^{\text{one-hot}} = \delta_y$ with a softened label distribution:

\begin{equation}
    y^{\text{LS}} = (1 - \epsilon) \delta_y + \frac{\epsilon}{K}
\end{equation}
where $0 < \epsilon < 1$ is a predefined smoothness hyper-parameter. By interpolating the one-hot label with a uniform distribution, label smoothing discourages over-confident predictions and prevents the classification loss from collapsing too quickly. As a result, it not only alleviates the loss scale imbalance between tasks but also improves generalization under distribution shifts. Accordingly, the classification loss with label smoothing is defined as:

\begin{equation}
    \mathcal{L}_{\text{CE-LS}} = - \sum_{i=1}^{K} y^{\text{LS}}_i \log p_i
\end{equation}

Finally, the overall optimization objective for domain alignment fine-tuning is given by:
\begin{equation}
    \label{eq:train_loss}
    \mathcal{L} = \mathcal{L}_{\text{CE-LS}} + \alpha \mathcal{L}_{\text{align}}
\end{equation}
where $\alpha > 0$ is a predefined weighting coefficient that controls the contribution of the representation alignment loss.

\subsection{Stable Model Ensembling}
\begin{figure}[t!]
    \centering
    \includegraphics[scale=0.6]{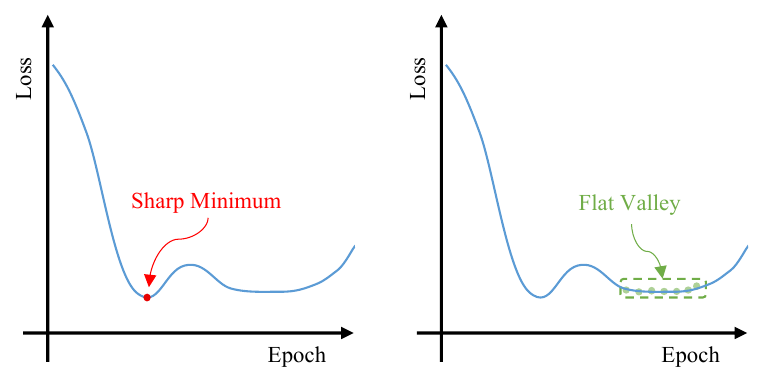}
    \caption{Illustration of a sharp minimum and a flat loss valley. A sharp minimum (left) is characterized by neighboring points exhibiting substantially higher average losses, whereas a flat loss valley (right) contains a local minimum together with a set of neighboring epochs whose losses remain comparable.}
    \label{fig:loss-illustration}
\end{figure}

During inference, the standard practice is to select the model checkpoint that achieves the lowest validation loss or highest validation accuracy during training. However, under a distribution shift, the validation and test data often follow different feature distributions. In such cases, a single validation-optimal checkpoint is prone to lie in a sharp minimum of the loss landscape, which typically exhibits poor generalization to unseen domains. 
In contrast, prior studies~\cite{izmailov2018averaging, cha2021swad, arpit2022ensemble} have shown that ensembling model weights located within a flat loss valley leads to substantially better generalization. An illustrative comparison between sharp minima and flat valleys is shown in \Cref{fig:loss-illustration}. 

Following this principle, \name{} adopts a weight ensembling strategy to obtain a merged model with more stable performance under distribution shifts. The overall procedure consists of three steps: (1) computing the validation loss at the end of each training epoch; (2) identifying a flat valley on the validation loss trajectory; and (3) averaging the model weights within the identified valley. As the first step is straightforward, we focus on valley identification and model averaging.

\emphtitle{Loss Valley Searching.} 
Intuitively, a sharp minimum is characterized by neighboring points with significantly higher average loss, whereas a flat valley contains a local minimum along with a range of neighboring epochs whose losses remain close. Following SWAD~\cite{cha2021swad}, we adopt an online heuristic to identify a flat valley from the validation loss trace. 

Specifically, we define the starting epoch of the flat valley, denoted as $t_{\text{s}}$, as the earliest epoch from which the validation loss does not decrease for $N_{\text{s}}$ consecutive epochs. The ending epoch, denoted as $t_{\text{e}}$, is defined as the earliest epoch after $t_{\text{s}}$ from which the validation loss consistently exceeds a threshold $\gamma$ for $N_{\text{e}}$ consecutive epochs. Formally, $t_{\text{s}}$ and $t_{\text{e}}$ are determined as:

\begin{equation}
    \label{eq:ts_te}
    \begin{aligned}
        t_{\text{s}} &= \min \left\{i \ \middle| \  \min_{0 \le j \le  N_{\text{s}} - 1} e_{i + j} = e_i \ ,\  1 \le i \le T_{\text{tr}} - N_{\text{s}} \right\} \\
        t_{\text{e}} &= \min \left\{i \ \middle| \  \min_{0 \le j \le  N_{\text{e}} - 1} e_{i + j} > \gamma \ ,\ t_{\text{s}} \le i \le T_{\text{tr}} - N_{\text{e}} \right\} \\
        \gamma &= \frac{r}{N_{\text{s}}} \sum_{i=0}^{N_{\text{s}}-1} e_{i + t_{\text{s}}}
    \end{aligned}
\end{equation}
where $e_i$ denotes the validation loss of the NTC model at the $i$-th epoch (computed using \Cref{eq:ce}), $T_{\text{tr}}$ is the total number of training epochs, and $r > 0$ is a predefined tolerance rate.

\emphtitle{Dynamic Model Averaging.}
After identifying the flat valley, a straightforward ensembling strategy is to average all model weights within the valley with equal weights~\cite{izmailov2018averaging, cha2021swad, arpit2022ensemble}. However, the heuristic nature of valley detection does not guarantee the exclusion of unstable model states associated with sharp loss spikes. Consequently, uniform averaging may be adversely affected by such unstable checkpoints.

To mitigate this issue, we introduce a dynamic weighting scheme that assigns lower importance to model states with higher validation loss. Specifically, the unnormalized and normalized weights for the model parameters at epoch $i$ are defined as:

\begin{equation}
    \begin{aligned}
        \tilde{w}_i &= \exp\left( \frac{e_{t_{\text{s}}} / e_i}{\tau} \right), \text{ for } t_{\text{s}} \le i \le t_{\text{e}} \\
        w_i &= \frac{\tilde{w}_i}{\sum_{j=t_{\text{s}}}^{t_{\text{e}}} w_j}, \text{ for } t_{\text{s}} \le i \le t_{\text{e}}
    \end{aligned}
\end{equation}
where $0 < \tau < 1$ is a temperature hyper-parameter. Using the validation loss at $t_{\text{s}}$, which corresponds to a local minimum sustained over at least $N_{\text{s}}$ epochs, as a reference point, model checkpoints with higher loss values receive smaller weights, while more stable checkpoints are emphasized. This mechanism effectively suppresses the influence of unstable model states on the final ensemble. 

Finally, the merged model parameters are obtained as a weighted average:
\begin{equation}
    \theta^{*} = \sum_{i = t_{\text{s}}}^{t_{\text{e}}} w_i \cdot \theta_i
\end{equation}
where $\theta_i$ denotes the model parameters after the $i$-th training epoch.

\begin{algorithm}[t!]
\caption{Training Procedure of \name{}}
\label{alg:training}
\begin{algorithmic}[1]
\Require Training data $D_{\text{tr}}$, validation data $D_{\text{va}}$, initial model parameters $\theta_0$, learning rate $\eta$, tolerance ratio $r$, convergence patience $N_{\text{s}}$, overfit patience $N_{\text{e}}$, total number of training epochs $T_{\text{tr}}$, temperature $\tau$
\Ensure Averaged model parameters $\theta^{*}$ from epochs $t_{\text{s}}$ to $t_{\text{e}}$

\State Initialize convergence queue $Q_{\text{c}} \leftarrow \varnothing$ with capacity $N_{\text{s}}$
\State Initialize overfit queue $Q_{\text{o}} \leftarrow \varnothing$ with capacity $N_{\text{e}}$
\State Initialize convergence epoch $t_{\text{s}} \leftarrow \varnothing$, ensembled parameters $\theta^{*} \leftarrow \varnothing$, cumulative weight $w_{\text{cum}} \leftarrow 0$, convergence loss $e_{t_\text{s}} \leftarrow 0$, overfit threshold $\gamma \leftarrow \varnothing$

\For{$i \leftarrow 1$ \textbf{to} $T_{\text{tr}}$}
    \State Compute training loss $\mathcal{L}(\theta_{i-1})$ on $D_{\text{tr}}$ using \cref{eq:train_loss} \label{line:train_loss}
    \State $\theta_i \leftarrow \theta_{i-1} - \eta \nabla \mathcal{L}(\theta_{i-1})$
    \State Compute validation loss $e_i$ on $D_{\text{va}}$ using \cref{eq:ce}
    \State Enqueue $(\theta_i, e_i)$ into $Q_{\text{c}}$ and $Q_{\text{o}}$ \label{line:queue_push}

    \If{$\gamma \neq \varnothing$ \textbf{and} all losses in $Q_{\text{o}}$ exceed $\gamma$} \Comment{detect overfitting} \label{line:overfit_start}
        \State \textbf{break} \label{line:overfit_end}

    \ElsIf{$Q_{\text{c}}$ is full \textbf{and} the loss at its head is minimal} \Comment{detect convergence} \label{line:converge_start}
        \State $t_{\text{s}} \leftarrow$ epoch index of the head of $Q_{\text{c}}$
        \State $e_{t_\text{s}} \leftarrow$ loss of the head of $Q_{\text{c}}$
        \State $\gamma \leftarrow r \times$ mean loss of $Q_{\text{c}}$
        \For{each $(\theta, e)$ in the first $N_{\text{s}} - N_{\text{e}}$ items of $Q_{\text{c}}$}
            \State $\theta^{*}, w_{\text{cum}} \leftarrow \Call{Update}{\theta^{*}, e_{t_\text{s}}, \theta, e, w_{\text{cum}}}$
        \EndFor \label{line:converge_end}

    \ElsIf{$t_{\text{s}} \neq \varnothing$} \Comment{converged but not yet overfitted} \label{line:update_start}
        \State Extract $(\theta, e)$ from the head of $Q_{\text{o}}$
        \State $\theta^{*}, w_{\text{cum}}  \leftarrow \Call{Update}{\theta^{*}, e_{t_\text{s}}, \theta, e, w_{\text{cum}}}$
    \EndIf \label{line:update_end}
\EndFor

\State \Return $\theta^{*}$

\Procedure{Update}{$\theta^{*}, e_{t_\text{s}}, \theta, e, w_{\text{cum}}$} \label{line:procedure_start}
    \State $\tilde{w} \leftarrow \exp((e_{t_{\text{s}}} / e)/\tau)$
    \State $\beta \leftarrow \tilde{w}/(w_{\text{cum}} + \tilde{w})$
    \State $\theta^{*} \leftarrow (1 - \beta)\theta^{*} + \beta \theta$
    \State $w_{\text{cum}} \leftarrow w_{\text{cum}} + \tilde{w}$
    \State \Return $\theta^{*}, w_{\text{cum}}$
\EndProcedure \label{line:procedure_end}

\end{algorithmic}
\end{algorithm}

\emphtitle{Joint Online Optimization.}
To avoid storing model checkpoints at every training epoch, \name{} maintains a limited number of temporary model states using sliding windows and performs model ensembling in an online fashion during training.
The complete procedure is summarized in \Cref{alg:training}. Specifically, Lines~\ref{line:train_loss}–\ref{line:queue_push} update the model parameters using the training data and record each updated model state along with its validation loss in the converge queue $Q_{\text{c}}$ and the overfit queue $Q_{\text{o}}$, respectively. These queues have fixed capacities of $N_{\text{s}}$ and $N_{\text{e}}$ and serve as sliding windows over recent training epochs.
Lines~\ref{line:overfit_start}–\ref{line:overfit_end} terminate the training process once the overfit epoch $t_{\text{e}}$, as defined in \Cref{eq:ts_te}, is detected. 
Lines~\ref{line:converge_start}–\ref{line:converge_end} identify the converge epoch $t_{\text{s}}$ (also defined in \Cref{eq:ts_te}) and initiate model ensembling by merging model states.
Once $t_{\text{s}}$ is determined but $t_{\text{e}}$ has not yet been reached, Lines~\ref{line:update_start}–\ref{line:update_end} continue to incorporate subsequent model states into the ensemble in an online manner.

Through this joint online optimization process, \name{} efficiently constructs a robust merged model without requiring specific traffic feature modalities or incurring non-constant additional computational overhead.

\section{Evaluation}
In this section, we evaluate the proposed training framework on three public datasets and compare its performance with seven state-of-the-art robust training methods.

\subsection{Experimental Setup}
\emphtitle{Datasets.}
We select three public network traffic datasets for evaluation, each of which exhibits a specific type of distribution shift.

\begin{itemize}[left=0pt, labelsep=1em, itemsep=0pt, topsep=0pt]
    \item \emph{CipherSpectrum}~\cite{wickramasinghe2025sok} contains encrypted network traffic collected by visiting 40 websites among the top 2000 domains listed on Cloudflare Radar, using both Firefox and Chromium browsers. For each website, traffic is collected iteratively under three TLS cipher suites (TLS-AES-128-GCM-SHA256, TLS-AES-256-GCM-SHA384, and TLS-CHACHA20-POLY1305-SHA256), with 1000 TLS 1.3 sessions recorded per suite. We treat each cipher suite as an individual domain and use this dataset to evaluate website classification performance under different encryption schemes.
    \item \emph{NUDT-Mobile}~\cite{zhao2024large} contains encrypted network traffic collected from 224 volunteers using at least nine smartphone brands and five operating system versions: Android 6, 7, 8, 9, and 10. Since Android 6 and 7 are largely outdated and account for only a small portion of the data, we merge them into a single OS version. Moreover, we retain only applications supported across all four OS versions. We treat each OS version as an individual domain and use this dataset to evaluate mobile application classification performance across different operating systems.
    \item \emph{CICIDS2017}~\cite{sharafaldin2018toward} contains benign traffic and several common, up-to-date attack types generated in a testbed network consisting of approximately 20 hosts. Attack traffic is generated over four days, with the dominant attacks of each day including brute-force, DoS, infiltration, and DDoS attacks, respectively. We treat each collection day as an individual domain and use this dataset to evaluate attack detection performance under different attack behaviors.\footnote{Since the attack behaviors differ substantially across days, using data from a single day leads to overly challenging classification tasks. To construct more tractable shifted datasets, we mix 0.5\% attack traffic from the subsequent day into each domain.}
\end{itemize}

\begin{table*}[ht]
    \footnotesize
    \centering
    \begin{tabular}{c|c|c|c|c}
        \toprule
         Dataset & Distribution Shift Type & Domains & \# Classes & \# Flows \\ \midrule
         CipherSpectrum & Encryption Scheme Shift & Aes128, Aes256, Chacha20 & 41 & 123000 \\ \midrule
         NUDT-Mobile & Collection Device Shift & Android6-7, Android8, Android9, Android10 & 16 & 50888 \\ \midrule
         CICIDS2017 & Attack Behavior Shift & Tuesday, Wednesday, Thursday, Friday & 2 & 40000 \\ \bottomrule
    \end{tabular}
    \caption{Summary of the evaluation datasets.}
    \label{tab:dataset}
\end{table*}

Key information regarding the distribution shift type, number of domains, classes, and traffic flows for the three datasets is summarized in \Cref{tab:dataset}. To reduce computational overhead and accelerate evaluation, we randomly sample at most $n$ traffic flows per class per domain, where $n=1000$ for CipherSpectrum and NUDT-Mobile, and $n=5000$ for CICIDS2017.

To fairly evaluate NTC model performance under distribution shifts, we conduct $k$-fold cross-domain evaluation on a dataset with $k$ domains. In each fold, data from one domain are held out for testing, while data from the remaining $k-1$ domains are randomly split into training and validation sets with a ratio of 9:1. Final results are reported by averaging performance across all $k$ folds.

For evaluation under i.i.d. settings, data from each domain are first randomly divided into training, validation, and test sets with an 8:1:1 ratio. The resulting splits from all domains within the same dataset are then merged to form the final training, validation, and test sets, respectively.

\emphtitle{NTC Models.}
We evaluate the following state-of-the-art supervised NTC models under distribution shifts.

\begin{itemize}[left=0pt, labelsep=1em, itemsep=0pt, topsep=0pt]
    \item \emph{NetMamba+}~\cite{wang2026netmamba+} is a pre-trained NTC model that processes traffic features from raw bytes, packet size sequences, and packet interval sequences. Its encoder consists of three embedding layers followed by four stacked Mamba blocks, and its classifier is a fully connected layer.

    \item \emph{NetTrans+}~\cite{wang2026netmamba+} is derived from NetMamba+. It replaces the Mamba blocks with four stacked Transformer blocks and adopts a different embedding dimension, while keeping other components identical to NetMamba+.
\end{itemize}
Both models have demonstrated strong performance across a wide range of NTC tasks under i.i.d. settings.

\emphtitle{Comparison Baselines.}
We first compare our framework with three robust NTC training methods.
\emph{MetaTraffic}~\cite{qing2025training} improves generalization to unseen domains through meta-learning and is applicable to NTC models with arbitrary input modalities.
In contrast, \emph{Rosetta}~\cite{xie2023rosetta} enhances generalization by augmenting packet size traces using TCP-related heuristics and BYOL-based contrastive learning, and is therefore restricted to models that process size sequences.
Similarly, \emph{NetAugment}~\cite{bahramali2023realistic} improves generalization by augmenting Tor traffic traces and applying SimCLR-based contrastive learning.

In addition, we compare \name{} with general domain generalization approaches.
\emph{MMD}~\cite{li2018domainmmd} aligns feature distributions across domains by minimizing pairwise distances in reproducing kernel Hilbert spaces. \emph{CDANN}~\cite{li2018domain} suppresses domain-specific information via adversarial training. \emph{SagNet}~\cite{nam2021reducing} separates content and style representations by replacing feature statistics. \emph{SWAD}~\cite{cha2021swad} improves generalization by uniformly averaging model weights within flat loss regions.

\emphtitle{Implementation Details.}
For NetMamba+ and NetTrans+, we adopt publicly available pre-trained checkpoints trained on the Browser~\cite{van2020flowprint} and Kitsune~\cite{mirsky2018kitsune} datasets. Input features are constructed as follows: for byte features, we extract the first 80 header bytes (excluding IP addresses and ports) and 80 payload bytes from the first 10 packets; for packet size and interval traces, we use the first 20 packets.

During fine-tuning, the batch size per domain is set to 64. For baseline methods, we use the default configurations reported in the corresponding papers or public implementations.

Hyperparameters for domain alignment fine-tuning and stable model ensembling are set as follows: label smoothing $\epsilon=0.1$, alignment loss coefficient $\alpha=0.5$, convergence patience $N_{\text{s}}=10$, overfitting patience $N_{\text{e}}=5$, tolerance ratio $r=1.1$, temperature $\tau=0.01$, and learning rate $\eta=2.0 \times 10^{-3}$.

All experiments are implemented in PyTorch 2.1.1 and conducted on a Ubuntu 22.04 server equipped with an Intel(R) Xeon(R) Gold 6240C CPU @ 2.60GHz and four NVIDIA A100 GPUs (40GB each).

\emphtitle{Evaluation Metrics.}
We evaluate NTC performance using two widely adopted metrics: \emph{accuracy (Acc)} and \emph{weighted F1 score (F1)}. Let $TP_c$, $FP_c$, $TN_c$, and $FN_c$ denote the numbers of true positives, false positives, true negatives, and false negatives for class $c$, respectively. Precision and recall are defined as $Pr_c = \frac{TP_c}{TP_c + FP_c}$ and $Rec_c = \frac{TP_c}{TP_c + FN_c}$, and let $n_c = TP_c + FN_c$ denote the number of samples in class $c$. The metrics are computed as:
\begin{equation}
    \begin{aligned}
        Acc &= \frac{\sum_c (TP_c + TN_c)}{\sum_c (TP_c + TN_c + FP_c + FN_c)}, \\
        F1 &= \sum_{c} \frac{n_c}{\sum_i n_i} \cdot \frac{2 \cdot Pr_c \cdot Rec_c}{Pr_c + Rec_c}.
    \end{aligned}
\end{equation}

For a dataset with $k$ domains, the final reported metrics are obtained by averaging over $k$ folds:
$Acc_{\text{avg}} = \frac{1}{k}\sum_{i=1}^{k} Acc_i$ and
$F1_{\text{avg}} = \frac{1}{k}\sum_{i=1}^{k} F1_i$.

\subsection{Performance under Distribution Shifts \label{sec:eval_ood}}

\begin{table*}[ht]
    \footnotesize
    \centering
    \begin{tabular}{c|c|cc|cc|cc}
         \toprule
         \multicolumn{2}{c|}{Training Framework} & \multicolumn{2}{c|}{CipherSpectrum} & \multicolumn{2}{c|}{NUDT-Mobile} & \multicolumn{2}{c}{CICIDS2017} \\ \midrule
         Type & Method & $Acc_{\text{avg}}$ & $F1_{\text{avg}}$ & $Acc_{\text{avg}}$ & $F1_{\text{avg}}$ & $Acc_{\text{avg}}$ & $F1_{\text{avg}}$ \\ \midrule
         Original & Standard & 0.6809$_{\scriptscriptstyle \pm 0.2391}$ & 0.6850$_{\scriptscriptstyle \pm 0.2258}$ & 0.6545$_{\scriptscriptstyle \pm 0.0632}$ & 0.6500$_{\scriptscriptstyle \pm 0.0672}$ & 0.9281$_{\scriptscriptstyle \pm 0.1090}$ & 0.9247$_{\scriptscriptstyle \pm 0.1158}$ \\ \midrule
         \multirow{4}{*}{General} & MMD~\cite{li2018domainmmd} & 0.6957$_{\scriptscriptstyle \pm 0.1924}$ & 0.6954$_{\scriptscriptstyle \pm 0.1822}$ & 0.6591$_{\scriptscriptstyle \pm 0.0663}$ & 0.6542$_{\scriptscriptstyle \pm 0.0707}$ & 0.9353$_{\scriptscriptstyle \pm 0.0813}$ & 0.9337$_{\scriptscriptstyle \pm 0.0842}$ \\ \cmidrule(lr){2-8}
          & CDANN~\cite{li2018domain} & 0.7017$_{\scriptscriptstyle \pm 0.2193}$ & 0.6977$_{\scriptscriptstyle \pm 0.2202}$ & 0.6597$_{\scriptscriptstyle \pm 0.0610}$ & 0.6547$_{\scriptscriptstyle \pm 0.0645}$ & 0.9528$_{\scriptscriptstyle \pm 0.0536}$ & 0.9524$_{\scriptscriptstyle \pm 0.0545}$ \\ \cmidrule(lr){2-8}
          & SagNet~\cite{nam2021reducing} & 0.6802$_{\scriptscriptstyle \pm 0.2152}$ & 0.6745$_{\scriptscriptstyle \pm 0.2131}$ & 0.6555$_{\scriptscriptstyle \pm 0.0611}$ & 0.6499$_{\scriptscriptstyle \pm 0.0620}$ & 0.9494$_{\scriptscriptstyle \pm 0.0464}$ & 0.9489$_{\scriptscriptstyle \pm 0.0472}$ \\ \cmidrule(lr){2-8}
          & SWAD~\cite{cha2021swad} & 0.6821$_{\scriptscriptstyle \pm 0.2085}$ & 0.6840$_{\scriptscriptstyle \pm 0.1953}$ & 0.6556$_{\scriptscriptstyle \pm 0.0653}$ & 0.6499$_{\scriptscriptstyle \pm 0.0690}$ & 0.9598$_{\scriptscriptstyle \pm 0.0587}$ & 0.9593$_{\scriptscriptstyle \pm 0.0597}$ \\ \midrule
         \multirow{3}{*}{NTC} & MetaTraffic~\cite{qing2025training} & 0.6942$_{\scriptscriptstyle \pm 0.2611}$ & 0.6961$_{\scriptscriptstyle \pm 0.2528}$ & 0.6636$_{\scriptscriptstyle \pm 0.0585}$ & 0.6601$_{\scriptscriptstyle \pm 0.0621}$ & \textbf{0.9640}$_{\scriptscriptstyle \pm 0.0486}$ & \textbf{0.9637}$_{\scriptscriptstyle \pm 0.0492}$ \\ \cmidrule(lr){2-8}
          & Rosetta~\cite{xie2023rosetta} & \textbf{0.7058}$_{\scriptscriptstyle \pm 0.2225}$ & 0.7020$_{\scriptscriptstyle \pm 0.2274}$ & 0.5936$_{\scriptscriptstyle \pm 0.0757}$ & 0.5891$_{\scriptscriptstyle \pm 0.0792}$ & 0.9092$_{\scriptscriptstyle \pm 0.0937}$ & 0.9060$_{\scriptscriptstyle \pm 0.0997}$ \\ \cmidrule(lr){2-8}
          & NetAugment~\cite{bahramali2023realistic} & 0.6772$_{\scriptscriptstyle \pm 0.2719}$ & 0.6724$_{\scriptscriptstyle \pm 0.2775}$ & 0.6548$_{\scriptscriptstyle \pm 0.0603}$ & 0.6510$_{\scriptscriptstyle \pm 0.0609}$ & 0.9537$_{\scriptscriptstyle \pm 0.0437}$ & 0.9534$_{\scriptscriptstyle \pm 0.0441}$ \\ \midrule
         \textbf{Ours} & \name{} & 0.7058$_{\scriptscriptstyle \pm 0.2031}$ & \textbf{0.7144}$_{\scriptscriptstyle \pm 0.1852}$ & \textbf{0.6751}$_{\scriptscriptstyle \pm 0.0593}$ & \textbf{0.6693}$_{\scriptscriptstyle \pm 0.0611}$ & 0.9451$_{\scriptscriptstyle \pm 0.0828}$ & 0.9437$_{\scriptscriptstyle \pm 0.0857}$ \\
         \bottomrule
    \end{tabular}
    \caption{NTC performance of different robust training frameworks on \textbf{NetMamba+}~\cite{wang2026netmamba+} under distribution shifts. }
    \label{tab:eval_ood_netmamba}
\end{table*}

\begin{table*}[ht]
    \footnotesize
    \centering
    \begin{tabular}{c|c|cc|cc|cc}
         \toprule
         \multicolumn{2}{c|}{Training Framework} & \multicolumn{2}{c|}{CipherSpectrum} & \multicolumn{2}{c|}{NUDT-Mobile} & \multicolumn{2}{c}{CICIDS2017} \\ \midrule
         Type & Method & $Acc_{\text{avg}}$ & $F1_{\text{avg}}$ & $Acc_{\text{avg}}$ & $F1_{\text{avg}}$ & $Acc_{\text{avg}}$ & $F1_{\text{avg}}$ \\ \midrule
         Original & Standard & 0.6863$_{\scriptscriptstyle \pm 0.2396}$ & 0.6886$_{\scriptscriptstyle \pm 0.2287}$ & 0.6766$_{\scriptscriptstyle \pm 0.0789}$ & 0.6704$_{\scriptscriptstyle \pm 0.0802}$ & 0.9645$_{\scriptscriptstyle \pm 0.0474}$ & 0.9642$_{\scriptscriptstyle \pm 0.0479}$ \\ \midrule
         \multirow{4}{*}{General} & MMD~\cite{li2018domainmmd} & 0.6531$_{\scriptscriptstyle \pm 0.2879}$ & 0.6592$_{\scriptscriptstyle \pm 0.2698}$ & 0.6754$_{\scriptscriptstyle \pm 0.0877}$ & 0.6699$_{\scriptscriptstyle \pm 0.0882}$ & 0.9744$_{\scriptscriptstyle \pm 0.0271}$ & 0.9743$_{\scriptscriptstyle \pm 0.0272}$ \\ \cmidrule(lr){2-8}
          & CDANN~\cite{li2018domain} & 0.6820$_{\scriptscriptstyle \pm 0.2621}$ & 0.6861$_{\scriptscriptstyle \pm 0.2494}$ & 0.6651$_{\scriptscriptstyle \pm 0.0717}$ & 0.6615$_{\scriptscriptstyle \pm 0.0738}$ & 0.9858$_{\scriptscriptstyle \pm 0.0085}$ & 0.9858$_{\scriptscriptstyle \pm 0.0085}$ \\ \cmidrule(lr){2-8}
          & SagNet~\cite{nam2021reducing} & 0.5946$_{\scriptscriptstyle \pm 0.3731}$ & 0.5952$_{\scriptscriptstyle \pm 0.3612}$ & 0.6747$_{\scriptscriptstyle \pm 0.0688}$ & 0.6704$_{\scriptscriptstyle \pm 0.0701}$ & 0.9497$_{\scriptscriptstyle \pm 0.0694}$ & 0.9488$_{\scriptscriptstyle \pm 0.0713}$ \\ \cmidrule(lr){2-8}
          & SWAD~\cite{cha2021swad} & 0.6811$_{\scriptscriptstyle \pm 0.2600}$ & 0.6867$_{\scriptscriptstyle \pm 0.2428}$ & 0.6824$_{\scriptscriptstyle \pm 0.0754}$ & 0.6766$_{\scriptscriptstyle \pm 0.0755}$ & 0.9747$_{\scriptscriptstyle \pm 0.0296}$ & 0.9746$_{\scriptscriptstyle \pm 0.0297}$ \\ \midrule
         \multirow{3}{*}{NTC} & MetaTraffic~\cite{qing2025training} & 0.6849$_{\scriptscriptstyle \pm 0.2359}$ & 0.6755$_{\scriptscriptstyle \pm 0.2460}$ & 0.6684$_{\scriptscriptstyle \pm 0.0798}$ & 0.6655$_{\scriptscriptstyle \pm 0.0753}$ & 0.9510$_{\scriptscriptstyle \pm 0.0644}$ & 0.9502$_{\scriptscriptstyle \pm 0.0659}$ \\ \cmidrule(lr){2-8}
          & Rosetta~\cite{xie2023rosetta} & 0.6812$_{\scriptscriptstyle \pm 0.1832}$ & 0.6764$_{\scriptscriptstyle \pm 0.1837}$ & 0.6623$_{\scriptscriptstyle \pm 0.0662}$ & 0.6613$_{\scriptscriptstyle \pm 0.0652}$ & 0.9829$_{\scriptscriptstyle \pm 0.0064}$ & 0.9829$_{\scriptscriptstyle \pm 0.0064}$ \\ \cmidrule(lr){2-8}
          & NetAugment~\cite{bahramali2023realistic} & 0.6501$_{\scriptscriptstyle \pm 0.2883}$ & 0.6489$_{\scriptscriptstyle \pm 0.2822}$ & 0.6672$_{\scriptscriptstyle \pm 0.0776}$ & 0.6646$_{\scriptscriptstyle \pm 0.0768}$ & 0.9834$_{\scriptscriptstyle \pm 0.0116}$ & 0.9834$_{\scriptscriptstyle \pm 0.0116}$ \\ \midrule
         \textbf{Ours} & \name{} & \textbf{0.7069}$_{\scriptscriptstyle \pm 0.2727}$ & \textbf{0.7122}$_{\scriptscriptstyle \pm 0.2583}$ & \textbf{0.6843}$_{\scriptscriptstyle \pm 0.0776}$ & \textbf{0.6786}$_{\scriptscriptstyle \pm 0.0786}$ & \textbf{0.9863}$_{\scriptscriptstyle \pm 0.0092}$ & \textbf{0.9863}$_{\scriptscriptstyle \pm 0.0092}$ \\
         \bottomrule
    \end{tabular}
    \caption{NTC performance of different robust training frameworks on \textbf{NetTrans+}~\cite{wang2026netmamba+} under distribution shifts. }
    \label{tab:eval_ood_nettrans}
\end{table*}

In this section, we evaluate the performance of \name{} alongside seven robust training baselines on two representative multimodal NTC models under distribution shifts. Each NTC model is fine-tuned separately using its original training strategy (denoted as \emph{Standard}), our framework, and each baseline method, and is then evaluated on the corresponding test data. The averaged results across all dataset–model–training framework combinations are summarized in \Cref{tab:eval_ood_netmamba,tab:eval_ood_nettrans}.

\emphtitle{Encryption Scheme Shift.}
We employ the CipherSpectrum dataset to simulate distribution shifts induced by different encryption schemes. The dataset is partitioned into three domains, namely Aes128, Aes256, and Chacha20, where each encryption scheme substantially alters traffic plaintexts and packet size traces. We conduct three-fold cross-domain evaluations, holding out one encryption scheme as the test domain in each fold. The averaged results are reported in \Cref{tab:eval_ood_netmamba,tab:eval_ood_nettrans}. As shown, \name{} improves $Acc_{\text{avg}}$ and $F1_{\text{avg}}$ of the two NTC models (NetMamba+ and NetTrans+) by an average of 0.0227 and 0.0265, respectively. In comparison, the best-performing baseline (CDANN) achieves smaller average gains of 0.0082 in $Acc_{\text{avg}}$ and 0.0051 in $F1_{\text{avg}}$.

\emphtitle{Collection Device Shift.}
We use the NUDT-Mobile dataset to evaluate robustness against distribution shifts caused by heterogeneous mobile devices during traffic collection. The dataset is divided into four domains corresponding to Android6–7, Android8, Android9, and Android10. Variations in operating system versions affect traffic characteristics through differences in protocol implementations, request ordering, and underlying network stacks. We perform four-fold evaluations by alternately designating each domain as the test set. The results in \Cref{tab:eval_ood_netmamba,tab:eval_ood_nettrans} show that \name{} improves $Acc_{\text{avg}}$ and $F1_{\text{avg}}$ by an average of 0.0142 and 0.0137 across the two NTC models, respectively. In contrast, the strongest baseline (SWAD) yields only marginal improvements of 0.0035 in $Acc_{\text{avg}}$ and 0.0030 in $F1_{\text{avg}}$.

\emphtitle{Attack Behavior Shift.}
To assess robustness under shifts in attack behaviors, we adopt the CICIDS2017 dataset, which is split into four domains corresponding to traffic collected on Tuesday, Wednesday, Thursday, and Friday. Each day contains distinct attack types that are absent from other days, resulting in pronounced behavioral shifts. We conduct four-fold evaluations, holding out one day as the test domain in each fold. As reported in \Cref{tab:eval_ood_netmamba,tab:eval_ood_nettrans}, \name{} improves $Acc_{\text{avg}}$ and $F1_{\text{avg}}$ of the two NTC models by an average of 0.0194 and 0.0206, respectively. Although the best baseline (CDANN) achieves slightly larger gains (0.0230 in $Acc_{\text{avg}}$ and 0.0247 in $F1_{\text{avg}}$), its improvements are less consistent across datasets and models.

\begin{figure}[tbp]
    \centering
    \begin{subfigure}[t]{\columnwidth}
        \centering
        \includegraphics[width=\linewidth]{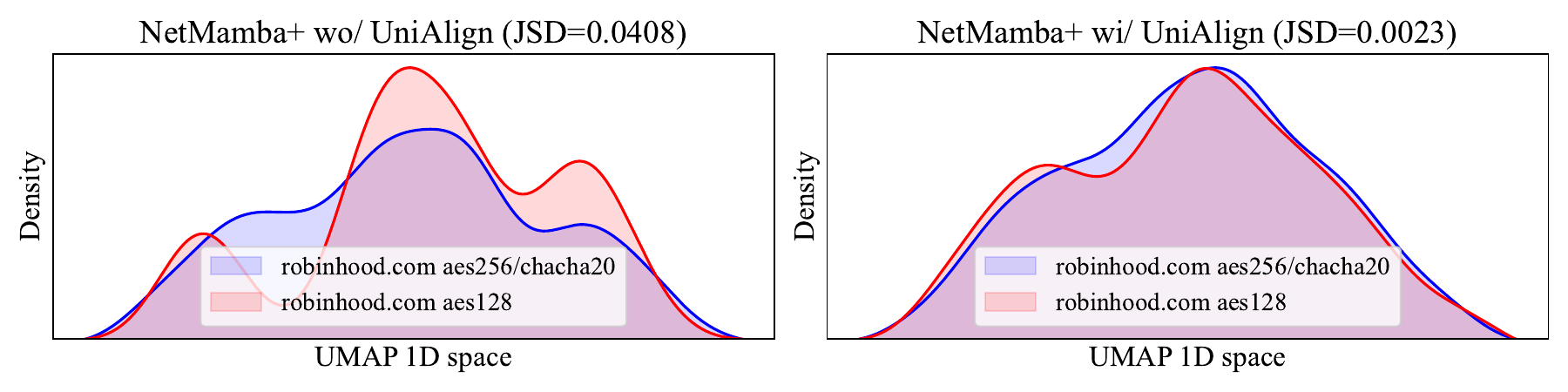}
    \end{subfigure}
    \begin{subfigure}[t]{\columnwidth}
        \centering
        \includegraphics[width=\linewidth]{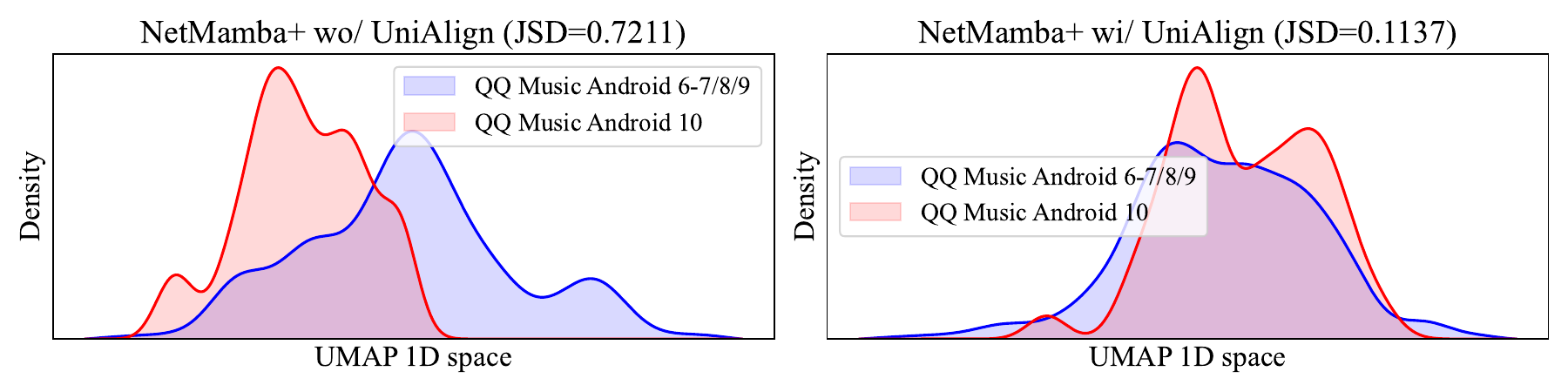}
    \end{subfigure}
    \begin{subfigure}[t]{\columnwidth}
        \centering
        \includegraphics[width=\linewidth]{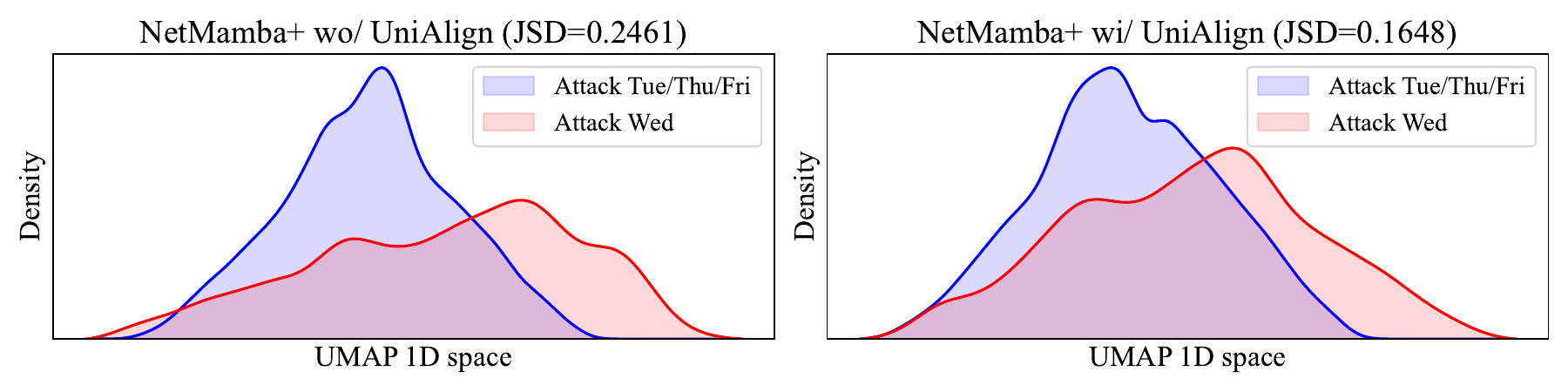}
    \end{subfigure}
    \caption{Visualization of feature representations produced by NetMamba+ fine-tuned without (left column) and with (right column) \name{} under distribution shifts. The top row shows the density of traffic from the \emph{robinhood.com} website in the CipherSpectrum dataset. The middle row corresponds to the density of the \emph{QQ Music} application in the NUDT-Mobile dataset. The bottom row presents the density of \emph{Attack} traffic in the CICIDS2017 dataset.}
    \label{fig:kde}
\end{figure}

Overall, our framework consistently enhances the robustness of both NTC models across all evaluated distribution shift scenarios. On average, \name{} achieves gains of 0.0188 (a relative improvement of 2.51\%) in $Acc_{\text{avg}}$ and 0.0203 (2.71\% relative) in $F1_{\text{avg}}$. In contrast, the best general-purpose baseline (CDANN) yields average improvements of 0.0094 (1.06\%) and 0.0092 (1.02\%), while the strongest NTC-specific baseline (MetaTraffic) achieves only 0.0059 (0.73\%) and 0.0047 (0.55\%) improvements in $Acc_{\text{avg}}$ and $F1_{\text{avg}}$, respectively.

Compared with general-purpose domain generalization methods, \name{} outperforms feature disentanglement approaches such as CDANN and SagNet by avoiding explicit domain label discrimination, which is particularly challenging given the high variability and ambiguity inherent in network traffic domains. Representation alignment methods (e.g., MMD) and model ensembling techniques (e.g., SWAD) exhibit limited effectiveness and stability when applied in isolation. Among NTC-specific baselines, MetaTraffic relies on meta-learning to simulate distribution shifts through multiple sub-tasks, which constrains its ability to capture diverse real-world shifts and introduces additional training overhead that scales with the number of sub-tasks. Rosetta and NetAugment, which depend on heuristic-based packet size augmentation, struggle to model complex and heterogeneous shift mechanisms, often resulting in degraded performance compared to standard training.

Our framework addresses these limitations by (1) improving fine-tuning through carefully designed domain alignment objectives and softened classification targets, and (2) enhancing inference robustness via stable model checkpoint averaging. To further illustrate the effectiveness of \name{}, we visualize feature representations generated by NetMamba+ under the three types of distribution shifts. Specifically, for encryption scheme shifts, NetMamba+ is trained on Aes256 and Chacha20 domains of CipherSpectrum; for collection device shifts, it is trained on Android6–7, Android8, and Android9 domains of NUDT-Mobile; and for attack behavior shifts, it is trained on Tuesday, Thursday, and Friday domains of CICIDS2017. For each scenario, we select traffic samples from a specific category, project their representations into one-dimensional space using UMAP~\cite{mcinnes2018umap}, estimate their probability density functions via kernel density estimation, and quantify distribution discrepancies between training and test domains using Jensen–Shannon divergence (JSD), as shown in \Cref{fig:kde}.

The results indicate that, under standard training, feature representations of traffic belonging to the same category but originating from different domains exhibit substantial divergence. In contrast, after applying \name{}, representations from training and unseen test domains become significantly more aligned. Quantitatively, the ratio of JSD without \name{} to that with \name{} ranges from 1.45 to 12.94, further confirming the effectiveness of \name{} in producing stable and domain-robust feature representations under distribution shifts.

\subsection{Performance under I.I.D. Settings \label{sec:eval_iid}}

\begin{table*}[ht]
    \footnotesize
    \centering
    \begin{tabular}{c|c|cc|cc|cc|cc|cc|cc}
         \toprule
         \multicolumn{2}{c|}{\multirow{2}{*}{Training Framework}} & \multicolumn{6}{c|}{NetMamba+~\cite{wang2026netmamba+}} & \multicolumn{6}{c}{NetTrans+~\cite{wang2026netmamba+}} \\ \cmidrule(lr){3-14}
         \multicolumn{2}{c|}{} & \multicolumn{2}{c|}{CipherSpectrum} & \multicolumn{2}{c|}{NUDT-Mobile} & \multicolumn{2}{c|}{CICIDS2017} & \multicolumn{2}{c|}{CipherSpectrum} & \multicolumn{2}{c|}{NUDT-Mobile} & \multicolumn{2}{c}{CICIDS2017} \\ \midrule
         Type & Method & $Acc$ & $F1$ & $Acc$ & $F1$ & $Acc$ & $F1$ & $Acc$ & $F1$ & $Acc$ & $F1$ & $Acc$ & $F1$ \\ \midrule

         Original & Standard & 0.9652 & 0.9654 & 0.8231 & 0.8232 & \textbf{0.9995} & \textbf{0.9995} & 0.9875 & 0.9875 & 0.8756 & 0.8758 & 0.9992 & 0.9992 \\ \midrule
         \multirow{4}{*}{General} & MMD~\cite{li2018domainmmd} & 0.9655 & 0.9656 & 0.8187 & 0.8194 & \textbf{0.9995} & \textbf{0.9995} & \textbf{0.9892} & \textbf{0.9893} & 0.8789 & 0.8792 & \textbf{0.9997} & \textbf{0.9997} \\ \cmidrule(lr){2-14}
         &  CDANN~\cite{li2018domain} & 0.9667 & 0.9670 & 0.8084 & 0.8090 & 0.9985 & 0.9985 & 0.9871 & 0.9871 & 0.8670 & 0.8670 & 0.9995 & 0.9995 \\ \cmidrule(lr){2-14}
         &  SagNet~\cite{nam2021reducing} & 0.9643 & 0.9641 & 0.8357 & 0.8359 & \textbf{0.9995} & \textbf{0.9995} & 0.9854 & 0.9854 & 0.8605 & 0.8608 & 0.9995 & 0.9995 \\ \cmidrule(lr){2-14}
         &  SWAD~\cite{cha2021swad} & 0.9698 & 0.9700 & 0.7999 & 0.8000 & \textbf{0.9995} & \textbf{0.9995} & 0.9881 & 0.9881 & 0.8657 & 0.8661 & 0.9995 & 0.9995 \\ \midrule
         \multirow{3}{*}{NTC} & MetaTraffic~\cite{qing2025training} & \textbf{0.9774} & \textbf{0.9775} & 0.8201 & 0.8215 & 0.9982 & 0.9982 & 0.9715 & 0.9717 & 0.8684 & 0.8686 & \textbf{0.9997} & \textbf{0.9997} \\ \cmidrule(lr){2-14}
         & Rosetta~\cite{xie2023rosetta} & 0.9541 & 0.9546 & 0.7488 & 0.7490 & 0.9982 & 0.9982 & 0.9805 & 0.9806 & 0.8623 & 0.8626 & \textbf{0.9997} & \textbf{0.9997} \\ \cmidrule(lr){2-14}
         & NetAugment~\cite{bahramali2023realistic} & 0.9648 & 0.9649 & 0.8353 & 0.8352 & \textbf{0.9995} & \textbf{0.9995} & 0.9773 & 0.9775 & 0.8787 & 0.8790 & 0.9992 & 0.9992 \\ \midrule

         \textbf{Ours}& $\name{}$ & 0.9723 & 0.9725 & \textbf{0.8391} & \textbf{0.8393} & 0.9992 & 0.9992 & 0.9860 & 0.9860 & \textbf{0.8906} & \textbf{0.8907} & \textbf{0.9997} & \textbf{0.9997} \\
         \bottomrule
    \end{tabular}
    \caption{NTC performance of different robust training frameworks under i.i.d. settings.}
    \label{tab:eval_iid}
\end{table*}

We evaluate \name{} and seven robust training baselines on two representative multimodal NTC models under commonly assumed i.i.d. settings, with the goal of examining whether robust training affects standard NTC performance. For each dataset, traffic flows from all domains are randomly mixed and split into unified training, validation, and test sets with an 8:1:1 ratio. The resulting NTC performance under i.i.d. settings is summarized in \Cref{tab:eval_iid}.

Compared with the results obtained under distribution shifts (\Cref{tab:eval_ood_netmamba,tab:eval_ood_nettrans}), both NetMamba+ and NetTrans+ trained using the standard method exhibit substantial performance improvements across all three datasets. On average, the two models achieve increases of 0.2927 in $Acc$ and 0.2897 in $F1$ on the CipherSpectrum dataset, 0.1838 in $Acc$ and 0.1893 in $F1$ on the NUDT-Mobile dataset, and 0.0531 in $Acc$ and 0.0549 in $F1$ on the CICIDS2017 dataset. This pronounced performance gap highlights the significant impact of distribution shifts on NTC models and further underscores the necessity of mitigating performance degradation through robust training strategies.

Notably, even under i.i.d. settings, our framework continues to provide consistent performance gains. Across the two models and three datasets, \name{} improves performance by an average of 0.0062 in $Acc$ and 0.0061 in $F1$. In contrast, other robust training baselines either yield only marginal improvements (e.g., MMD and NetAugment) or even degrade performance under i.i.d. conditions. Specifically, the average accuracy improvement of all baselines ranges from -0.0046 to 0.0008, while the average $F1$ improvement ranges from -0.0046 to 0.0008.

The performance gains of \name{} can be attributed to two key factors. First, the softened classification target discourages over-confident predictions, thereby alleviating overfitting. Second, the stable model ensembling strategy preserves model states with better generalization capability and predictive accuracy. Together, these mechanisms enable \name{} to learn feature representations that are both stable and discriminative, resulting in improved NTC performance under both i.i.d. and distribution-shifted settings.

\subsection{Training Efficiency \label{sec:efficiency}}

\begin{figure}[tbp]
    \centering
    \includegraphics[width=\linewidth]{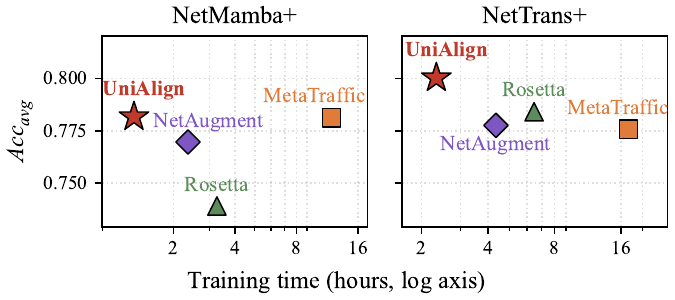}
    \caption{Average training time versus OOD accuracy of different training frameworks on NetMamba+ (left) and NetTrans+ (right). Training time is averaged across all OOD folds of the three datasets, measured on a dedicated single GPU. 
    }
    \label{fig:efficiency}
\end{figure}

We further compare the training cost of \name{} against three NTC-specific robust training baselines from \Cref{sec:eval_ood}—MetaTraffic~\cite{qing2025training}, Rosetta~\cite{xie2023rosetta}, and NetAugment~\cite{bahramali2023realistic}—on a dedicated single GPU. To ensure a fair comparison, all frameworks fine-tune the backbone for 120 epochs, and the contrastive pretraining stages of Rosetta and NetAugment use an additional 60 epochs. As shown in \Cref{fig:efficiency}, \name{} is the only framework on the Pareto frontier of both backbones: averaged across the two backbones, its training time is 53.9\% of NetAugment, 37.1\% of Rosetta, and only 12.4\% of MetaTraffic, while still attaining 1.25 percentage points higher $Acc_{\text{avg}}$ than the next-best framework.

This efficiency advantage stems from three differences in how baselines extend the standard fine-tuning pipeline. First, \name{} only adds an extra optimization objective during standard fine-tuning, without increasing the epoch complexity or altering the model architecture. Second, Rosetta and NetAugment require an additional self-supervised contrastive pretraining stage over augmented samples before fine-tuning, and Rosetta further inserts an extra TIE module into the model architecture during fine-tuning, raising both the schedule length and the per-step cost. Third, MetaTraffic's two-stage inner/outer optimization makes the per-epoch training step count scale linearly with the number of sub-tasks (4 on NUDT-Mobile and CICIDS2017, 3 on CipherSpectrum), substantially inflating its overall training time.

\subsection{Ablation Study \label{sec:ablation}}

\begin{table*}[ht]
    \footnotesize
    \centering
    \begin{tabular}{c|cc|cc|cc}
        \toprule
        \multirow{2}{*}{Training Setting} & \multicolumn{2}{c|}{CipherSpectrum} & \multicolumn{2}{c|}{NUDT-Mobile} & \multicolumn{2}{c}{CICIDS2017} \\ \cmidrule(lr){2-7}
         & $Acc_{\text{avg}}$ & $F1_{\text{avg}}$ & $Acc_{\text{avg}}$ & $F1_{\text{avg}}$ & $Acc_{\text{avg}}$ & $F1_{\text{avg}}$ \\ \midrule
        Full \name{} & 0.7058$_{\scriptscriptstyle \pm 0.2031}$ & 0.7144$_{\scriptscriptstyle \pm 0.1852}$ & \textbf{0.6751}$_{\scriptscriptstyle \pm 0.0593}$ & \textbf{0.6693}$_{\scriptscriptstyle \pm 0.0611}$ & 0.9451$_{\scriptscriptstyle \pm 0.0828}$ & 0.9437$_{\scriptscriptstyle \pm 0.0857}$ \\ \midrule
        wo/ DAF & \textbf{0.7120}$_{\scriptscriptstyle \pm 0.2210}$ & \textbf{0.7147}$_{\scriptscriptstyle \pm 0.2117}$ & 0.6577$_{\scriptscriptstyle \pm 0.0674}$ & 0.6524$_{\scriptscriptstyle \pm 0.0710}$ & \textbf{0.9651}$_{\scriptscriptstyle \pm 0.0488}$ & \textbf{0.9648}$_{\scriptscriptstyle \pm 0.0494}$ \\ \midrule
        wo/ SME & 0.7063$_{\scriptscriptstyle \pm 0.2061}$ & 0.7123$_{\scriptscriptstyle \pm 0.1936}$ & 0.6744$_{\scriptscriptstyle \pm 0.0586}$ & 0.6676$_{\scriptscriptstyle \pm 0.0603}$ & 0.9151$_{\scriptscriptstyle \pm 0.0594}$ & 0.9138$_{\scriptscriptstyle \pm 0.0614}$ \\
        \bottomrule
    \end{tabular}
    \caption{NTC performance with \name{} and its variants on \textbf{NetMamba+} under distribution shifts. }
    \label{tab:eval_ablation_netmamba}
\end{table*}

\begin{table*}[ht]
    \footnotesize
    \centering
    \begin{tabular}{c|cc|cc|cc}
        \toprule
        \multirow{2}{*}{Training Setting} & \multicolumn{2}{c|}{CipherSpectrum} & \multicolumn{2}{c|}{NUDT-Mobile} & \multicolumn{2}{c}{CICIDS2017} \\ \cmidrule(lr){2-7}
         & $Acc_{\text{avg}}$ & $F1_{\text{avg}}$ & $Acc_{\text{avg}}$ & $F1_{\text{avg}}$ & $Acc_{\text{avg}}$ & $F1_{\text{avg}}$ \\ \midrule
        Full \name{} & \textbf{0.7069}$_{\scriptscriptstyle \pm 0.2727}$ & \textbf{0.7122}$_{\scriptscriptstyle \pm 0.2583}$ & 0.6843$_{\scriptscriptstyle \pm 0.0776}$ & \textbf{0.6786}$_{\scriptscriptstyle \pm 0.0786}$ & \textbf{0.9863}$_{\scriptscriptstyle \pm 0.0092}$ & \textbf{0.9863}$_{\scriptscriptstyle \pm 0.0092}$ \\ \midrule
        wo/ DAF & 0.6585$_{\scriptscriptstyle \pm 0.3073}$ & 0.6692$_{\scriptscriptstyle \pm 0.2841}$ & 0.6814$_{\scriptscriptstyle \pm 0.0738}$ & 0.6755$_{\scriptscriptstyle \pm 0.0747}$ & 0.9499$_{\scriptscriptstyle \pm 0.0477}$ & 0.9495$_{\scriptscriptstyle \pm 0.0481}$ \\ \midrule
        wo/ SME & 0.6916$_{\scriptscriptstyle \pm 0.3180}$ & 0.7013$_{\scriptscriptstyle \pm 0.2971}$ & \textbf{0.6847}$_{\scriptscriptstyle \pm 0.0720}$ & 0.6779$_{\scriptscriptstyle \pm 0.0745}$ & 0.9296$_{\scriptscriptstyle \pm 0.0837}$ & 0.9277$_{\scriptscriptstyle \pm 0.0873}$ \\
        \bottomrule
    \end{tabular}
    \caption{NTC performance with \name{} and its variants on \textbf{NetTrans+} under distribution shifts. }
    \label{tab:eval_ablation_nettrans}
\end{table*}

In this section, we conduct an ablation study to examine the individual effectiveness of the two core modules in \name{} under distribution shifts. To this end, we construct two variants of our framework by removing one module at a time, evaluate their performance, and compare them with the full \name{}. Specifically, to assess the contribution of the domain alignment fine-tuning module, we design a variant denoted as \emph{wo/ DAF}, which fine-tunes the NTC model using standard cross-entropy loss without the domain alignment target or the softened classification target. To evaluate the stable model ensembling module, we create another variant, \emph{wo/ SME}, which directly selects the single checkpoint achieving the highest validation accuracy for inference, instead of performing model ensembling. All variants are evaluated under the same distribution shift settings described in \Cref{sec:eval_ood}, and the results are reported in \Cref{tab:eval_ablation_netmamba,tab:eval_ablation_nettrans}.

As shown in the table, removing either module leads to a noticeable degradation in performance. Overall, compared with the full \name{}, the averaged $Acc_{\text{avg}}$ and $F1_{\text{avg}}$ of NetMamba+ and NetTrans+ decrease by 0.0131 and 0.0131, respectively, when trained with \emph{wo/ DAF}, and by 0.0169 and 0.0173 when trained with \emph{wo/ SME}. These results demonstrate that both modules play a crucial role in improving the stability and robustness of learned feature representations, thereby enabling \name{} to achieve superior NTC performance under distribution shifts.

\subsection{Deep Dive}

In this section, we conduct an in-depth comparison of the design choices in \name{} for both the domain alignment fine-tuning module and the stable model ensembling module. By contrasting our approach with several alternative designs, we further validate the effectiveness and rationality of our selected configurations.

\begin{figure}[tbp]
    \centering
    \includegraphics[width=\linewidth]{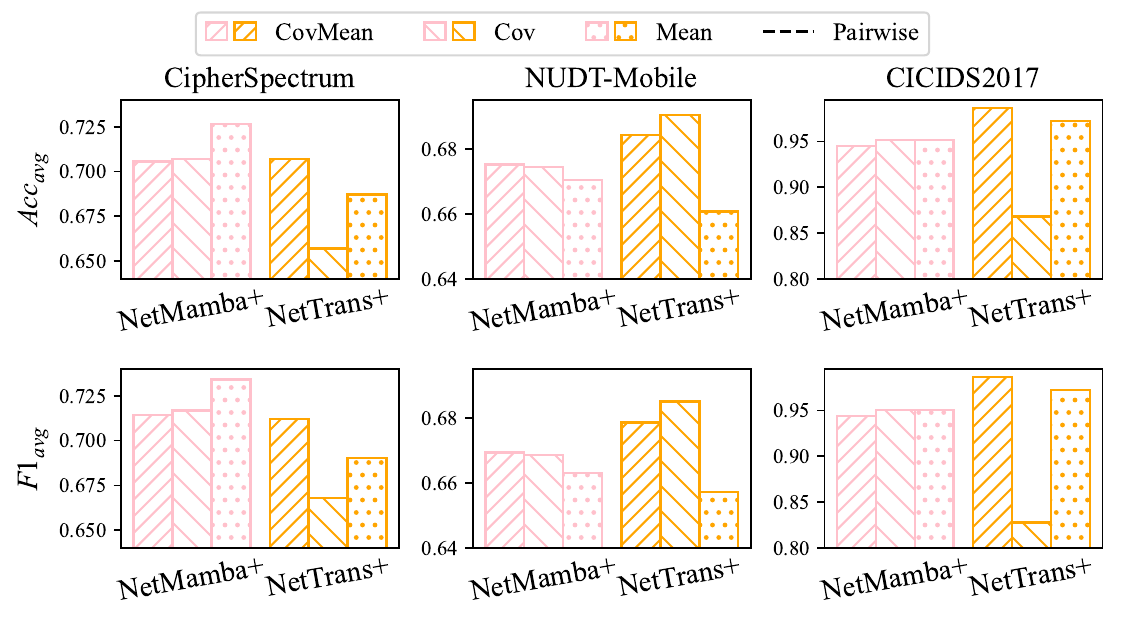}
    \caption{The impact of different distance metrics on NTC performance under distribution shifts.}
    \label{fig:dist_cmp}
\end{figure}

\emphtitle{Domain Alignment Target.}
To evaluate the effectiveness of our proposed alignment loss, which minimizes cross-domain distributional distances measured by sample means and covariances (\emph{CovMean}), we compare it against three alternative metrics. Specifically, \emph{Mean} considers only the distance between sample means, \emph{Cov} measures the distance between covariance matrices, and \emph{Pairwise} computes instance-level distances by accumulating distances over all sample pairs.

As shown in \Cref{fig:dist_cmp}, the instance-level \emph{Pairwise} metric yields the worst performance across all evaluation settings, with results falling far below the visible y-axis ranges. This inferior performance can be attributed to unstable distance estimations caused by outlier samples. In contrast, the three distribution-level metrics demonstrate more competitive performance, each excelling in certain evaluation combinations. When averaged across all evaluation combinations, \emph{CovMean} achieves the best overall performance, with $Acc_{\text{avg}} = 0.7839$ and $F1_{\text{avg}} = 0.7841$. Compared to \emph{CovMean}, \emph{Cov} lags behind by 0.0259 in $Acc_{\text{avg}}$ and 0.0313 in $F1_{\text{avg}}$, while \emph{Mean} trails by 0.0058 and 0.0062, respectively. The gap is even more pronounced for \emph{Pairwise}, which underperforms by 0.3806 in $Acc_{\text{avg}}$ and 0.4194 in $F1_{\text{avg}}$. These results highlight the effectiveness of our alignment target that jointly captures first-order and second-order statistics.

\begin{figure}[tbp]
    \centering
    \includegraphics[width=\linewidth]{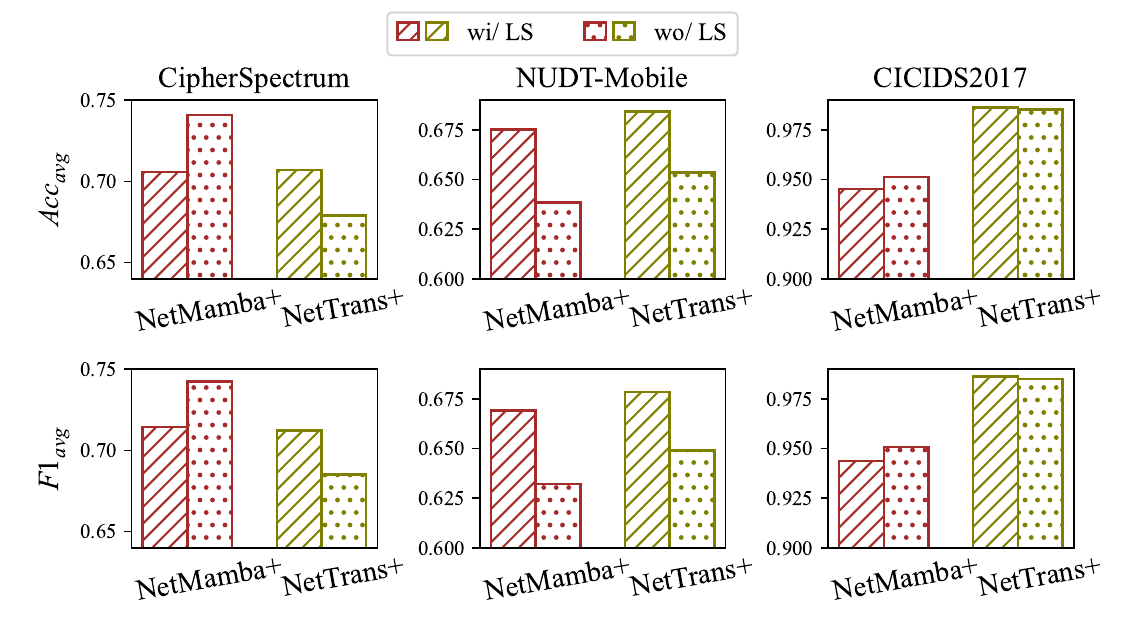}
    \caption{The impact of label smoothing on NTC performance under distribution shifts.}
    \label{fig:ls_performance}
\end{figure}

\emphtitle{Training Loss Balancing.}
To assess the effectiveness of our proposed training loss balancing strategy based on label smoothing, we compare the full framework equipped with label smoothing (\emph{wi/ LS}) against a variant that removes label smoothing from the classification loss (\emph{wo/ LS}). The resulting NTC performance under distribution shifts is shown in \Cref{fig:ls_performance}. Overall, NetTrans+ consistently benefits from label smoothing, whereas NetMamba+ experiences performance degradation on two datasets when label smoothing is applied. Nevertheless, when averaged across two models and three datasets, \emph{wi/ LS} improves upon \emph{wo/ LS} by 0.0092 in $Acc_{\text{avg}}$ and 0.0100 in $F1_{\text{avg}}$.

\begin{figure}[tbp]
    \centering
    \includegraphics[width=\linewidth]{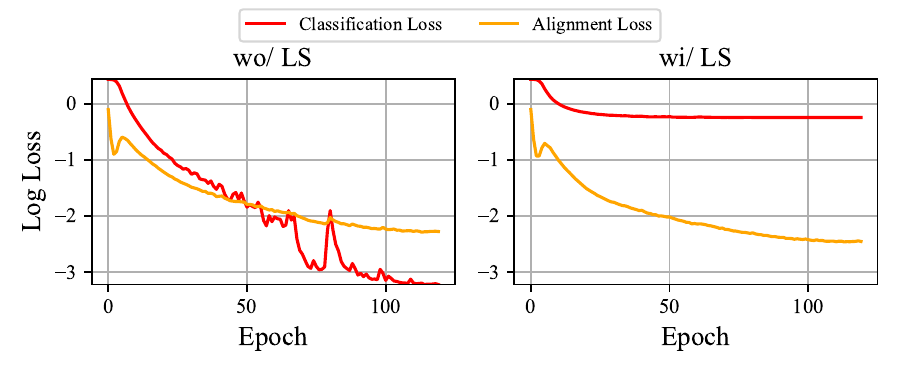}
    \caption{The impact of label smoothing on training loss scales under distribution shifts.}
    \label{fig:ls_loss}
\end{figure}

To further analyze the source of these improvements, we compare the training loss dynamics of NetTrans+ under \emph{wi/ LS} and \emph{wo/ LS} on the NUDT-Mobile dataset, with Android10 as the test domain. As illustrated in \Cref{fig:ls_loss}, when training without label smoothing, the classification loss (in log scale) decreases almost linearly with training epochs. After approximately half of the training process, the classification loss becomes an order of magnitude smaller than the alignment loss, causing the optimization objective to be dominated by alignment and ultimately impairing classification performance. In contrast, introducing label smoothing ensures that the classification loss consistently exceeds the alignment loss by at least one order of magnitude, thereby effectively preserving NTC performance.

\emphtitle{Model Ensembling Strategy.}
To evaluate the effectiveness of our proposed stable model ensembling (\emph{SME}), which averages checkpoints located within a flat validation loss region, we compare it with two alternative ensembling strategies. Specifically, \emph{Global} ensembling averages all checkpoints obtained at the end of each training epoch, regardless of validation performance, while \emph{Top-10} ensembling averages the ten checkpoints with the lowest validation losses. In addition, we include the single checkpoint with the highest validation accuracy (\emph{BestAcc}) as a non-ensemble baseline.

\begin{figure}[tbp]
    \centering
    \includegraphics[width=\linewidth]{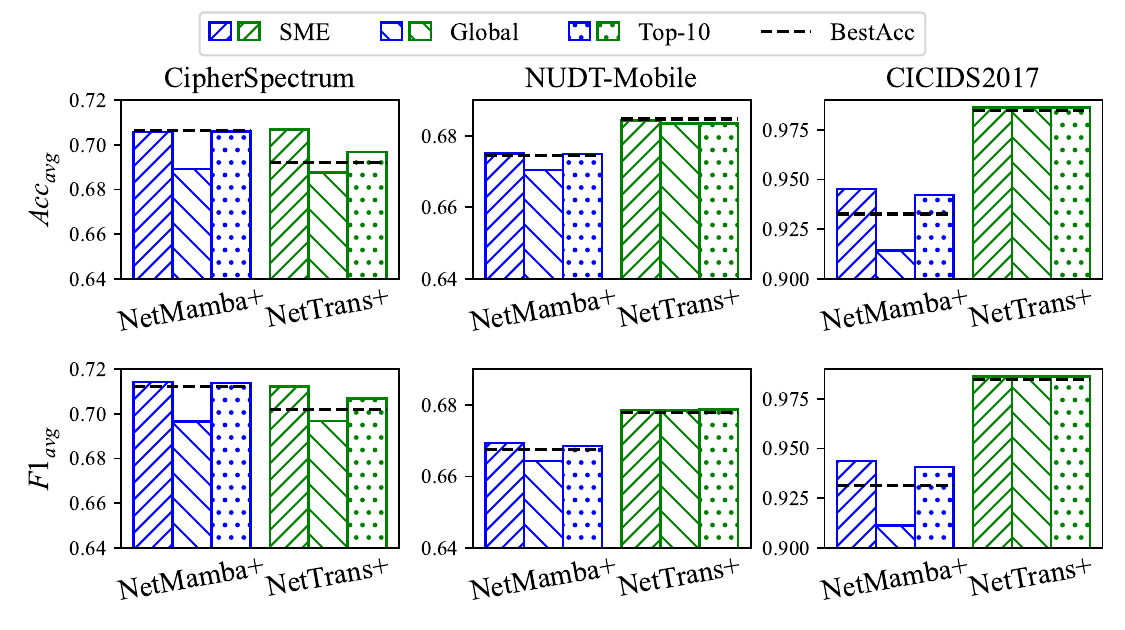}
    \caption{NTC performance of different ensembling variants under distribution shifts.}
    \label{fig:ensemble}
\end{figure}

As shown in \Cref{fig:ensemble}, both \emph{SME} and \emph{Top-10} consistently outperform the non-ensemble \emph{BestAcc} baseline across most evaluation settings, whereas the brute-force \emph{Global} ensembling method performs significantly worse than the other approaches. Quantitatively, \emph{SME} further improves upon \emph{Top-10} by 0.0021 in $Acc_{\text{avg}}$ and 0.0015 in $F1_{\text{avg}}$, averaged over two models and three datasets. These results confirm that \emph{SME} is more effective at identifying and aggregating stable model states, making it better suited for deployment under distribution shifts.

\subsection{Discussion \label{sec:discussion}}

\emphtitle{Training Data Collection.}
Our framework assumes that the training traffic data can be partitioned into multiple non-overlapping domains, where each domain corresponds to a specific network condition. In practice, such a partition can be guided by various factors, including collection time periods, end devices, geographic locations, or network environments. These factors naturally induce distribution shifts due to differences in transmission paths, link congestion levels, protocol implementations, and underlying network infrastructures.
The domain alignment module in our framework operates by minimizing discrepancies across training domains and therefore requires the number of source domains to satisfy $S \geq 2$. When this condition cannot be met (i.e., $S = 1$), the domain alignment loss degenerates to zero, and the framework effectively reduces to the \emph{wo/ DAF} variant described in \Cref{sec:ablation}. In this case, the average performance of our framework decreases by 0.0131 in $Acc_{\text{avg}}$ and 0.0131 in $F1_{\text{avg}}$ across two models and three datasets. Nevertheless, owing to the generalization benefits provided by the stable model ensembling module, the degraded framework still outperforms standard training by an average of 0.0056 in $Acc_{\text{avg}}$ and 0.0072 in $F1_{\text{avg}}$.
On the other hand, when multiple plausible criteria for domain partitioning are available, determining an optimal number of domains and an appropriate splitting strategy remains a non-trivial problem. We leave the systematic study of domain construction strategies for traffic data as future work.

\emphtitle{NTC Model Selection.}
Our framework is designed to improve the robustness of supervised deep learning–based NTC models under distribution shifts, irrespective of their underlying feature modalities. Due to page limitations and computational constraints, we evaluate our framework on two representative multimodal models that jointly leverage raw-byte-level and sequential features and adopt Transformer- or Mamba-based architectures.
In principle, both the domain alignment module and the stable model ensembling module are model-agnostic and should generalize to state-of-the-art NTC models that rely solely on sequential features (e.g., FSNet~\cite{liu2019fs}, FlowPic~\cite{shapira2019flowpic}) or raw bytes (e.g., YaTC~\cite{zhao2023yet}, ET-BERT~\cite{lin2022bert}).
To empirically validate this claim, we additionally train FSNet on the same three datasets using both the standard training procedure and our framework. Under standard training, FSNet achieves an average performance of 0.7414 in $Acc_{\text{avg}}$ and 0.7365 in $F1_{\text{avg}}$, while our framework improves these metrics by 0.0104 and 0.0100, respectively. This result further demonstrates the applicability of our framework to enhancing the generalization performance of existing NTC models.

\emphtitle{NTC Model Deployment.}
In real-world deployments, supervised NTC models inevitably encounter diverse forms of distribution shifts. Existing robust training frameworks, including ours, primarily address closed-set shift scenarios, where the traffic category set remains unchanged from training. However, open-set scenarios, in which previously unseen traffic categories emerge, are also common in operational networks. Such scenarios may arise from the deployment of new applications or services, as well as the appearance of novel attack tools.
While our framework is not designed to directly identify new traffic categories, the stable and discriminative feature representations it produces provide a favorable foundation for a wide range of model-agnostic out-of-distribution (OOD) detection techniques. These include methods based on prediction probabilities~\cite {hendrycks2016baseline, liu2020energy, liang2017enhancing, liu2023gen}, representation-space distances~\cite{sehwag2021ssd, sun2022out, park2023nearest}, and gradient-based measures~\cite{huang2021importance, chen2023gaia, behpour2023gradorth}.
Once OOD traffic is detected, active learning strategies~\cite{li2024survey} can be employed to acquire reliable labels, after which NTC models may be updated either incrementally through lifelong learning approaches~\cite{zhou2023deep} or retrained from scratch. A comprehensive evaluation of open-set robustness is beyond the scope of this work and is left for future investigation.

\section{Related Work}

\emphtitle{Deep Learning for NTC.}
Due to the superior modeling capabilities inherent in deep learning, a diverse range of deep neural networks~(DNNs) have been proposed to address NTC tasks. Early approaches leveraged various architectures to capture distinct traffic features.
For example, FlowPic~\cite{shapira2019flowpic} and mini-FlowPic~\cite{horowicz2022few} employ CNNs to capture traffic patterns represented within 2-dimensional histograms of flow data. FS-Net~\cite{liu2019fs} utilizes RNNs, specifically a GRU autoencoder, to process sequences of packet sizes. GraphDApp~\cite{shen2021accurate} uses GNNs to model traffic interaction graphs, while TFE-GNN~\cite{zhang2023tfe} leverages GNNs to learn correlations within byte-level traffic graphs. PERT~\cite{he2020pert} and MTT~\cite{zheng2022mtt} process raw traffic bytes using Transformer models, demonstrating an ability to capture long-range dependencies.

More recently, inspired by the huge breakthroughs in large language models, the NTC area has begun to embrace pre-trained models. For instance, models like ET-BERT~\cite{lin2022bert}, PTU~\cite{peng2024ptu}, TrafficFormer~\cite{zhou2025trafficformer}, and MM4flow~\cite{yang2025mm4flow} adapt the BERT paradigm. They achieve robust pre-training by splitting datagrams into tokens and learning token semantics through a masked prediction objective. FlowMAE~\cite{hang2023flow}, YaTC~\cite{zhao2023yet}, and NetMamba~\cite{wang2024netmamba} follow the MAE paradigm. They split flow bytes into consecutive patches and pre-train the model by reconstructing the masked patches. NetGPT~\cite{meng2023netgpt}, Lens~\cite{wang2024lens}, and PcapEncoder~\cite{zhao2025sweet} employ generative pre-trained models (e.g., GPT-2 and T5) to capture traffic patterns, often focusing on predicting the next byte or sequence.

However, all of above deep learning-based NTC models are currently evaluated under an independent and identically distributed (i.i.d.) setting. Consequently, they frequently suffer from significant performance degradation when deployed in real-world environments characterized by distribution shifts.

\emphtitle{Robust Training for NTC.}
To achieve robust performance under distribution shifts, recent methods have proposed both model-specific approaches, which are constrained to particular data acquisition scenarios or model architectures, and model-agnostic approaches, which can seamlessly adapt existing models.
Among model-specific methods, CD-Net~\cite{chen2025cd} retrains its DNN classifier using data collected from the new distribution while keeping the feature extractor fixed. FG-SAT~\cite{cui2025fg} filters stable graph features across different network environments, and ETooL~\cite{lin2025respond} leverages the inherent generalization ability of LLMs like Vicuna-7B-v1.5. In contrast, model-agnostic methods, such as NetAugment~\cite{bahramali2023realistic} and Rosetta~\cite{xie2023rosetta}, enhance the generalization of NTC models by augmenting packet traces with heuristic sequence modifications. Additionally, MetaTraffic~\cite{qing2025training} improves out-of-distribution (OOD) generalization abilities through meta learning, albeit at the cost of requiring significantly more training iterations.

In comparison to existing methods, our model-agnostic framework enhances the NTC performance of models based on any input features and does so without incurring additional training iterations beyond the standard training procedure.

\section{Conclusion}

In this paper, we propose \name{}, a novel model-agnostic training framework designed to improve the robustness of deep learning-based NTC models under distribution shifts. Our framework enhances NTC robustness by promoting the learning of stable feature representations during fine-tuning and by improving inference reliability through the averaging of stable model checkpoints. Compared with existing robust training baselines, \name{} can be seamlessly integrated into supervised NTC models without requiring specific feature modalities or incurring non-constant additional training costs. We evaluate the effectiveness of \name{} on three public datasets using two representative NTC models. Under distribution-shifted settings, \name{} improves average accuracy by 2.51\% and average F1 score by 2.71\% compared with standard training, and outperforms the strongest baseline by 1.45\% in accuracy and 1.69\% in F1 score, while requiring only 12.4\%--53.9\% of the training time of all NTC-specific baselines.


\bibliographystyle{IEEEtran}
\bibliography{reference}

\end{document}